\documentclass[journal]{IEEEtran}
\usepackage{jsen}
\usepackage{cite}
\usepackage{amsmath,amssymb,amsfonts}
\usepackage{algorithmic}
\usepackage{graphicx}
\usepackage{textcomp}
\usepackage{wrapfig}
\usepackage{booktabs}
\usepackage{orcidlink}
\usepackage{subfigure}
\usepackage{float}
\usepackage{tabularx} 
\usepackage{siunitx}
\usepackage{changes}
\usepackage{longtable}
\usepackage{xcolor}
\usepackage{algorithm}
\usepackage{array}
\usepackage{stfloats}
\usepackage{url}
\usepackage{verbatim}
\usepackage{dcolumn}
\usepackage{multirow}

\def\BibTeX{{\rm B\kern-.05em{\sc i\kern-.025em b}\kern-.08em
    T\kern-.1667em\lower.7ex\hbox{E}\kern-.125emX}}
\definecolor{abstractbg}{rgb}{0.89804,0.94510,0.83137}
\begin{document}
\title{MultiFormer: A Multi-Person Pose Estimation \\System Based on CSI and Attention Mechanism}
\author{Yanyi Qu$^{\orcidlink{0009-0003-6183-1502}}$, Haoyang Ma$^{\orcidlink{0009-0001-2207-169X}}$, and Wenhui Xiong$^{\orcidlink{0000-0003-2051-0457}}$
\thanks{Manuscript received May 28, 2025; revised [Month, Day], [Year]. This work was supported by the National Key Laboratory of Science and Technology on Communications, University of Electronic Science and Technology of China (UESTC).(Corresponding author: Wenhui Xiong)}
\thanks{Y. Qu and H. Ma are with the School of Information and Communication Engineering, University of Electronic Science and Technology of China, Chengdu 611731, China (e-mail: [2022010910012@std.uestc.edu.cn]; [2022010901019@std.uestc.edu.cn]).}
\thanks{W. Xiong is with the National Key Laboratory of Science and Technology on Communications, UESTC, Chengdu 611731, China (e-mail: whxiong@uestc.edu.cn).}}

\maketitle
\begin{abstract}
Human pose estimation based on Channel State Information (CSI) has emerged as a promising approach for non-intrusive and precise human activity monitoring, yet faces challenges including accurate multi-person pose recognition  and effective CSI feature learning. This paper presents MultiFormer, a wireless sensing system that accurately estimates human pose through CSI. The proposed system adopts a Transformer based time-frequency dual-token feature extractor with multi-head self-attention. This feature extractor is able to model inter-subcarrier correlations and temporal dependencies of the CSI. The extracted CSI features and the pose probability heatmaps are then fused by Multi-Stage Feature Fusion Network (MSFN) to enforce the anatomical constraints. Extensive experiments conducted on on the public MM-Fi dataset and our self-collected dataset show that the MultiFormer achieves higher accuracy over state-of-the-art approaches, especially for high-mobility keypoints (wrists, elbows) that are particularly difficult for previous methods to accurately estimate. 

\end{abstract}

\section{Introduction}
\label{sec:introduction}
\IEEEPARstart{H}{uman} pose estimation, a new human-machine interaction technique, aims to extract human activity information from sensor signals, thereby finding extensive applications in smart healthcare, smart homes, virtual reality, and motion analysis \cite{minhdangSensorbasedVisionbasedHuman2020,liuWirelessSensingHuman2020,pareekSurveyVideobasedHuman2021,guptaHumanActivityRecognition2022}. Existing human pose estimation methods can be categorized into two types: Vision-based and Non-Vision-based. Vision-based systems typically rely on optical sensors, e.g., RGB cameras \cite{cao_openpose_2021}, depth sensors \cite{DEPsensor}, or infrared arrays \cite{infrared} and computer vision algorithms to provide precise and high-resolution estimation results. However, their applications are limited by lighting conditions and privacy concerns \cite{pareekSurveyVideobasedHuman2021}\cite{cao_openpose_2021}\cite{chu_structured_2016,fang_rmpe_2017,ronchi_benchmarking_2017,song_thinslicing_2017,huang_confidencebased_2022}. Non-Vision-based methods, such as radio frequency (RF)-based sensing \cite{Radar1} addressed the limitations of vision-based approaches, and emerged as a promising approach for human pose estimation, since it doesn't require the person of interest to wear special sensors \cite{mcgrath_bodyworn_2020}\cite{niehorster_accuracy_2017}. Among RF-based solutions, radars offer certain advantages but are limited by the need for specialized equipment, which can be costly and complex to deploy. In contrast, WiFi-based pose estimation utilizes existing wireless infrastructure and enables low-cost implementation without requiring additional equipment \cite{chen_lowcost_2021}. 

Recent research on WiFi-based human pose estimation focuses on Channel State Information (CSI) which provides finer measurements of multi-path effects across subcarriers and antenna links \cite{yangRSSICSIIndoor2013}. Different human poses alter the multi-path environment, where WiFi signals propagate, differently. Thus, one can estimate human pose by capturing the variations of CSI.

CSI-based human pose estimation systems typically consist of three key modules: data preprocessing module, feature extraction module, and pose estimation module.

The data preprocessing module performs data sanitization and transformation of raw CSI into suitable inputs for neural networks. Inspired by the success of computer vision, many studies organize the CSI across different subcarriers, antennas, and time instants as a tensor, e.g., \cite{guAttentionBasedGestureRecognition2023,yang_Metafi_2022,zhouMetafiWiFiEnabledTransformerBased2023,jiang_3d_2020,zhouCSIFormerPayMore2022,wangCanWiFiEstimate2019}. However, this direct adaptation ignores the difference between CSI and visual signals. Unlike the spatially correlated RGB pixels, 
the temporal, spatial, and frequency
components of the CSI tensor are of different correlations. Therefore, applying CNN on such CSI tensors with sliding windows by assuming that adjacent areas of the tensor have similar features, lacks interpretability and often results in poor pose estimation results. 

The feature extraction module captures the feature representations of human pose information from CSI. Traditional CNN-based methods require multiple layers to achieve global perception. Researchers have explored various architectures ranging from 16-layer ResNet backbones \cite{wangCanWiFiEstimate2019}, adaptive convolution mechanisms \cite{zhouCHASensEndtoEndComprehensive2023}, to hybrid CNN-GRU frameworks \cite{wangPointSpace3D2021}. These CNN-based modules require a deep network of multiple layers for global feature extraction, which yields a high computation load and poor performance for complex pose estimation.

Recent studies show Transformer, as a long-range and global feature extractor, can effectively capture dependencies from the entire sequence. Many CSI-based Human Pose Estimation systems adopt Transformer as their feature extractor. Hybrid framework like Metafi++ \cite{zhouMetafiWiFiEnabledTransformerBased2023} keeps CNN components that constrain attention scope. CSI-Former \cite{zhouCSIFormerPayMore2022} uses a self-attention-based architecture that models inter-subcarrier correlations but overlooks temporal dependencies, limiting its ability to capture time-varying dynamics.

Pose estimation module estimates the coordinates of human pose keypoints by processing features extracted from CSI. Most existing methods can be divided into two categories: direct keypoint regression and pose adjacency matrix (PAM) regression.

The keypoints regression method used in Wi-Mose\cite{wangPointSpace3D2021} and Metafi \cite{yang_Metafi_2022}\cite{zhouMetafiWiFiEnabledTransformerBased2023} directly decoded the coordinates of human keypoints through multilayer perceptrons. These methods treat the estimation of human keypoint coordinates as independent regression tasks. Thus, Such design fails to capture the constraints between human keypoints. The PAM regression methods used in CSI-Former \cite{zhouCSIFormerPayMore2022} and 
WiSPPN\cite{wangCanWiFiEstimate2019} encode the relative position between keypoints for pose estimation. However, PAM-based approaches are limited to model pairwise spatial distances and fail to capture the more complex constraints between keypoints.

Direct keypoints regression and PAM regression approaches have fixed output dimensions, which restrict the pose estimation to single-person scenarios. To extend CSI-based pose sensing to multi-person scenarios, the heatmap-based method was proposed. It uses the heatmap of Part Confidence Maps (PCM) and Part Affinity Fields (PAF) to identify keypoints and matching algorithms to connect the keypoints across different individuals \cite{wang_personinwifi_2019}\cite{huang_crossmodal_2021}. However, this approach decodes heatmaps of all keypoints in parallel, which, similar to \cite{zhouCSIFormerPayMore2022}\cite{wangCanWiFiEstimate2019}, also ignores complex constraints between keypoints.

To address these problems, we propose the MultiFormer system. Unlike conventional CSI imagization techniques, MultiFormer uses the "Time-Frequency Dual-Dimensional Tokenization (TFDDT)" method for CSI preprocessing. This method utilizes a dual-token Transformer architecture with multi-head self-attention to extract features from CSI. TFDDT is able to capture both spectral correlations and temporal dynamics. Additionally, MultiFormer adopts a multi-stage heatmap-based estimation approach to enable global perception of human poses. By iteratively refining estimations of different stages, MultiFormer enhances estimation accuracy for multi-person scenarios. Extensive experiments conducted on both the public dataset, i.e., MM-Fi, and our self-collected multi-person dataset validate the effectiveness of our proposed method.

The contribution of this paper is summarized as follows:

\begin{enumerate}
    \item \textbf{Time-Frequency Dual-Dimensional Tokenization of CSI} We propose to transform the raw CSI into ``time tokens'' and ``frequency tokens''. Such design preserves the  continuity of the feature in both temporal and frequency domains, which avoids the distortions introduced by cross domain convolutions, which facilitates the enhancement of feature learning capability in later process.

    \item \textbf{Self-Attention Based CSI Feature extraction} Our model employs a multi-head self-attention mechanism to capture inter-subcarrier correlations and temporal dependencies of CSI data. This approach enhances the feature learning by focusing on the key subcarriers and the key intervals.
    
    \item \textbf{Multi-Stage Pose Estimation} We propose a Multi-Stage Feature Fusion Network (MSFN) that iteratively refines pose estimations through the fusion of CSI features and intermediate pose probability heatmaps (PCM/PAF). This iterative refinement ensues accurate, anatomically realistic estimation results.

\end{enumerate}

\section{Related Work}

\subsection{CSI Preprocessing}
The preprocessing of CSI includes methods such as direct CSI tensor reshaping, conversion to spectrograms, and tokenization followed by embedding. Gu et al.\cite{guAttentionBasedGestureRecognition2023} used the quotient of the CSI from two adjacent antennas of the same receiver to eliminate random phase offsets. Zhou et al.\cite{zhouMetafiWiFiEnabledTransformerBased2023} reshaped the CSI into a matrix, with the first dimension consisting of frequency information and the second dimension consisting of antenna and time information. Niu et al.\cite{niuCSIFHumanMotion2024} directly converted the CSI into a 2D spectrogram. Yan et al.\cite{yan_personinwifi_2024} proposed a tokenization method for CSI samples, where CSI of each transmit-receive antenna pair at each time sample is converted into a single token. 

Existing methods directly convert CSI into image-like tensors, which neglect the CSI' temporal and frequency domains characteristics. Our proposed Time-Frequency Dual-Dimensional Tokenization (TFDDT) addresses this by creating time and frequency token streams, which enable more effective feature extraction.

\subsection{Feature Extraction of CSI}
 Traditional methods, such as CNNs-based methods, often require multiple layers to achieve global perception. For example, Wang et al. \cite{wangCanWiFiEstimate2019} used a 16-layer ResNet to extract features from upsampled CSI. Zhou et al. \cite{zhouCHASensEndtoEndComprehensive2023} proposed an adaptive convolution method for feature extraction. Wang et al. \cite{wangPointSpace3D2021} used a 13-layer Residual Network to extract pose-related features from CSI. Tang et al. \cite{tangMDPoseHumanSkeletal2024} proposed a denoising network (FMNet) that extracts clean micro-Doppler features from noisy spectrograms and uses CNNs and RNNs to learn temporal-spatial dependencies. Nguyen et al.  \cite{nguyen_robust_2025} implemented efficient feature extraction from CSI using a dynamic convolutional neural network (SDy-CNN). 

Recently, Transformer has become an important framework in the fields of natural language processing (NLP) and computer vision (CV). Researchers have turned to Transformer-based architecture due to its long-range dependency capture capability. For instance,  Zhou et al. \cite{zhouCSIFormerPayMore2022} introduced the CSI-Former model, which combines attention mechanisms with traditional CNNs to process CSI. This model allows the network to focus on those CSI subcarriers that are more important for pose estimation, thereby improving the efficiency and accuracy of feature extraction. Zhou et al. 

\cite{zhouMetafiWiFiEnabledTransformerBased2023} proposed the Metafi++, which employs a CNN module to extract features from each pair of antennas, followed by a channel-wise self-attention mechanism to model interdependencies across different feature channels. Some studies inspired by Vision Transformers (ViT) have adopted the patch embedding idea \cite{yangWiTransformerNovelRobust2023}\cite{luoVisionTransformersHuman2024} to analyze CSI spectra for human activity recognition.

Existing CNN-based feature extraction methods often struggle with capturing long-range dependencies and require significant computational resources. While Transformer-based models can capture long-range dependencies, they tend to neglect temporal dynamics. Our method’s dual-path attention captures both spectral and temporal features with the same or fewer parameter count and computational cost, giving better efficiency and performance.

\subsection{Human Pose Estimation Techniques}
Accurately estimating the positions of skeletal keypoints is crucial in human pose estimation. Current methods for this task can be categorized into direct regression, PAM regression, and heatmap-based approaches. Wang et al. \cite{wangPointSpace3D2021} proposed the Wi-Mose model, which uses a series of perceptrons to directly regress the coordinates of each keypoint. Similarly, Metafi and Metafi++ \cite{yang_Metafi_2022}\cite{zhouMetafiWiFiEnabledTransformerBased2023} employ multiple layers of perceptrons to estimate keypoint coordinates. 
CSI-Former \cite{zhouCSIFormerPayMore2022} leverages the distance differences between keypoints to build a PAM, which captures the spatial relationships between keypoints. Wang et al. \cite{wang_personinwifi_2019} used PAFs to enable multi-person pose estimation. Huang et al. \cite{huang_crossmodal_2021} combined the OpenPose network \cite{cao_openpose_2021} with a custom CSI2Pose network to estimate PCMs and PAFs from CSI to determine the correct keypoint connections\cite{cao_openpose_2021}. They used PCMs and PAFs to achieve multi-person pose estimation.

Conventional single-stage pose estimation methods, i.e.,
keypoints regression, estimate the pose directly but fail to
model the inter-keypoints connections properly. While the
PAM-based models is able to capture the pair-wise links
between keypoints, the full-body coherence is still a challeng-
ing task. We address this task by cascading three heatmap-
based feature fusion stages by Pose-Attentive modules that re-
weight features spatially and channel-wise to give structurally
consistent results.

\begin{figure*}[!t]
\centering
\includegraphics[width=\textwidth]{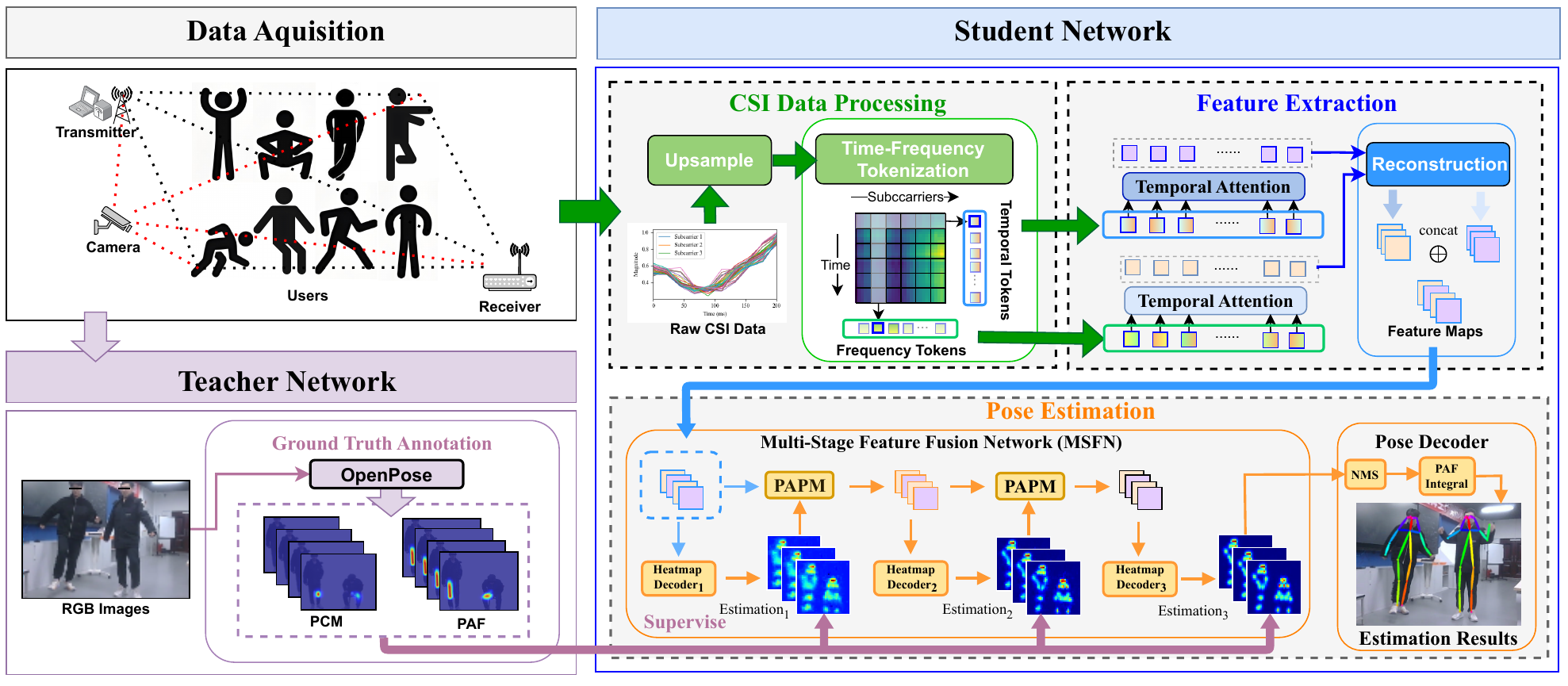}
\caption{Block Diagram Of MultiFormer}
\label{fig:OVO}
\end{figure*}

\section{System Architecture}
Our proposed MultiFormer system adopts a teacher-student network architecture. As depicted in Figure~\ref{fig:OVO}, the teacher network uses a vision-based sensor model to obtain the ground truth of the human pose to train the student network. The student network, after training, uses CSI to estimate human poses without the need of vision-based sensor input. 

For teacher network, we adopt the model used in OpenPose \cite{cao_openpose_2021} to estimate PCM and PAF as labels. PCM and PAF enable the teacher network to handle multi-person scenarios without pre-estimated bounding boxes. The student network in the MultiFormer system comprises three key modules, including the CSI Preprocessing module, the Feature Extraction module and the Multi-Stage Pose Estimation module.

\subsection{CSI Preprocessing}

We use the TFDDT method in MultiFormer to transform CSI into token sequences for the transformer-based network to extract the features. 

For devices compliant with the IEEE 802.11 standard, the physical layer employs Orthogonal Frequency-Division Multiplexing (OFDM) and multiple antenna techniques. OFDM technology transforms the wide band channel into multiple orthogonal sub-channels and converts the high-speed serial data flow into multiple lower-speed parallel data flows. The CSI of a MIMO-OFDM system can be expressed in matrix form as:

\begin{equation}
{\mathbf{H}_s(i)} = 
\begin{pmatrix}
h_s^{11}(i) & h_s^{12}(i) & \ldots & h_s^{1N_R}(i) \\
h_s^{21}(i) & h_s^{22}(i) & \ldots & h_s^{2N_R}(i) \\
\vdots & \vdots & \ddots & \vdots \\
h_s^{N_T1}(i) & h_s^{N_T2}(i) & \ldots & h_s^{N_TN_R}(i)
\end{pmatrix}
\end{equation}where $h_s^{mn}(i)$ is the complex channel response for the $s$-th subcarrier ($1 \leq s \leq N_S$) from the $m$-th transmit antenna ($1 \leq m \leq N_T$) to the $n$-th receive antenna ($1 \leq n \leq N_R$) at the $i$-th time instant.

\begin{figure}[h]
\centering
\includegraphics[width=3.5in]{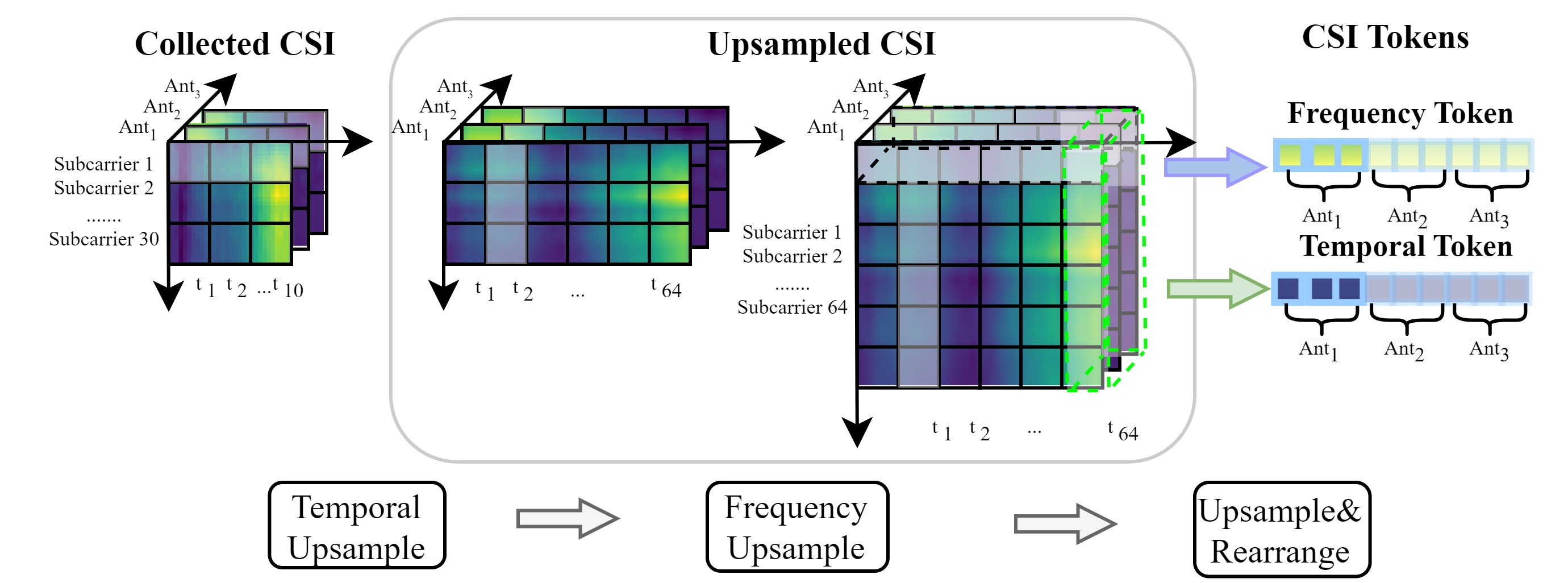}
\caption{Workflow of Time-Frequency Dual-Dimensional Tokenization (TFDDT)}
\label{fig:TFDDTOken}
\end{figure}

As illustrated in Fig.~\ref{fig:TFDDTOken}, TFDDT upsamples the raw CSI measurements along time (packet sequence) and frequency (OFDM subcarrier) dimensions. The upsampled data is then reorganized into Frequency Tokens and Temporal Tokens. The Frequency Token incorporates amplitude variations across time and antennas for a specific subcarrier, whereas the Temporal Token encapsulates the frequency-domain patterns across subcarriers and antennas at a particular sampling moment. 

Our experimental system is with 1 transmitter and 3 receivers, and the CSI is collected at the rate of 50 Hz. We upsample the collected CSI in time and frequency domain to match the input dimension of the attention network, i.e., from $10 \times 3 \times 30$ (package number $\times$ receiver antennas $\times$ subcarriers) to $64 \times 3 \times 64$.

The upsampling is achieved by inserting zeros along the time and frequency dimensions, followed by low-pass filtering. We show the time domain upsample as an example, and the upsample in the frequency domain can be done by following the similar method. The time domain upsample can be expressed as:

\begin{equation}
\widehat{h}_{s}^{mn} =  g_T \ast \left( \sum_{k=-\infty}^{\infty}  |h_{s}^{mn}(k)| \delta\left(i - kN\right) \right) 
\end{equation}where the $g_T$ denotes low-pass filters for temporal anti-aliasing, $\ast$ is the convolution operation, and the $N$ is the upsample ratio. It is worth noting that, in this work, we use CSI amplitude $|{h}_{s}^{mn}|$ to generate tokens since the amplitude captures the variations of CSI caused by different human poses. The upsampling operation in the frequency dimension is consistent with that in the temporal dimension.

After the upsampling in temporal and frequency domain, we group the upsampled CSI to form the Frequency Token and Time Token, denoted as \( F_j \) and \( T_i \). For the Frequency Token \( F_j \), the upsampled CSI is organized into \( N_S \) one-dimensional tokens of length \( M \times N_R \). In this work, $M=64$ is the number of upsampled CSI packets, $N_R = 3$ is the number of receiving antennas, and $N_S = 64$ is the dimension of upsampled subcarriers. Each Frequency Token represents the combination of CSI at different moments and different receiving antennas of a given subcarrier. The Frequency Token \( F_j \) at the \( j \)-th subcarrier is represented as:

\begin{equation}
\begin{split}
F_j = \big[ & |\widehat{h}_j^{11}(1)|, |\widehat{h}_j^{11}(2)|, \ldots, |\widehat{h}_j^{11}(M)|, \\
& |\widehat{h}_j^{12}(1)|, |\widehat{h}_j^{12}(2)|, \ldots, |\widehat{h}_j^{12}(M)|, \ldots, \\
& |\widehat{h}_j^{1N_R}(1)|, |\widehat{h}_j^{1N_R}(2)|, \ldots, |\widehat{h}_j^{1N_R}(M)| \big]^T
\end{split}
\end{equation}

Similarly, the Temporal Token \( T_i \) is generated by grouping CSI into \( M \) one-dimensional tokens of length \( N_S \times N_R \). Each time token is the combination of CSI from different subcarriers and receiving antennas at the same moment given by:

\begin{equation}
\begin{split}
T_i = \big[ & |\widehat{h}_1^{11}(i)|, |\widehat{h}_1^{12}(i)|, \ldots, |\widehat{h}_1^{1N_R}(i)|, \\
& |\widehat{h}_2^{11}(i)|, |\widehat{h}_2^{12}(i)|, \ldots, |\widehat{h}_2^{1N_R}(i)|, \ldots, \\
& |\widehat{h}_s^{11}(i)|, |\widehat{h}_s^{12}(i)|, \ldots, |\widehat{h}_s^{1N_R}(i)| \big]^T
\end{split}
\end{equation}

In this work, the TFDDT generates 64 frequency tokens and 64 temporal tokens, each with a length of \(1 \times 1296\). This design extracts information from both temporal and frequency dimensions through separate token streams. The learnable embeddings with random initialization, each of dimension \(1 \times 1296\), are added to each token to provide additional context and distinguishability. After adding the embeddings, the input sequence has a shape of \(64 \times 1296\), which balances computation load and pose estimation accuracy, enabling the model to capture fine-grained spatial patterns in CSI.

\begin{figure*}[!t]
\centering
\includegraphics[width=\textwidth]{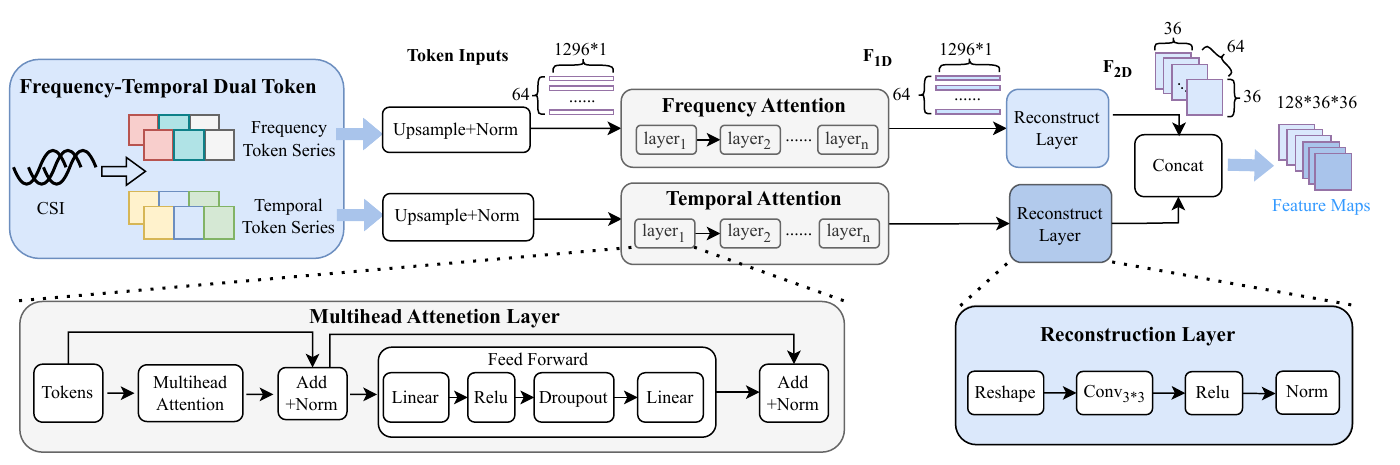}
\caption{Block diagram of multi-head self-attention feature extraction module}
\label{fig:FeatureEX}
\end{figure*}

\subsection{Feature Extraction}
After the tokenization of the raw CSI, two parallel multi-head self-attention modules with independent parameters are employed to process the token series and extract features. As shown in Fig.~\ref{fig:FeatureEX}, each path composites a normalization layer, a standard multi-head attention layer, and a reconstruction layer.

We use the Frequency Token sequence as an example to show the process of feature extraction, and the Temporal Token follows the similar approach.

\begin{figure*}[htbp]
  \centering
  \subfigure[\label{fig:push1}]
  {\includegraphics[width=1.7in]{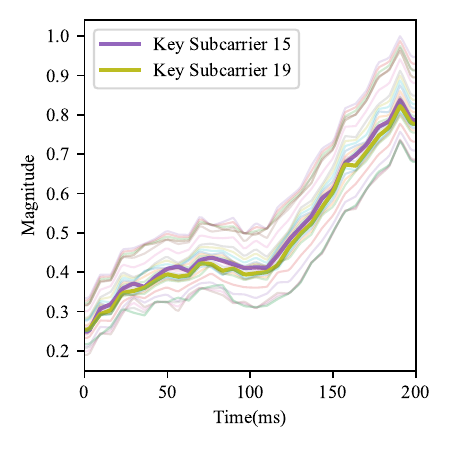}}%
  \hfill 
  \subfigure[\label{fig:push2}]
  {\includegraphics[width=1.7in]{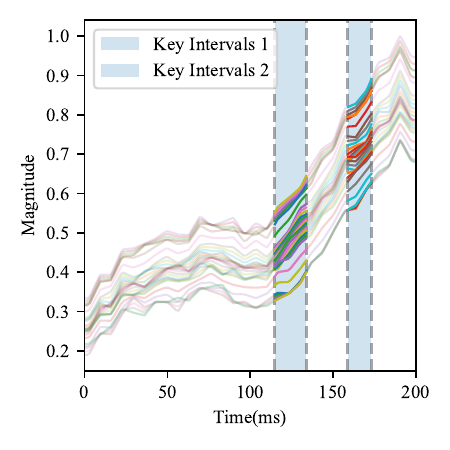}}%
  \hfill 
  \subfigure[\label{fig:bend1}]
  {\includegraphics[width=1.7in]{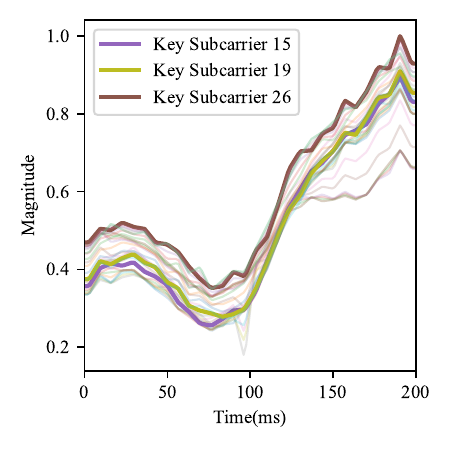}}%
  \hfill 
  \subfigure[\label{fig:bend2}]
  {\includegraphics[width=1.7in]{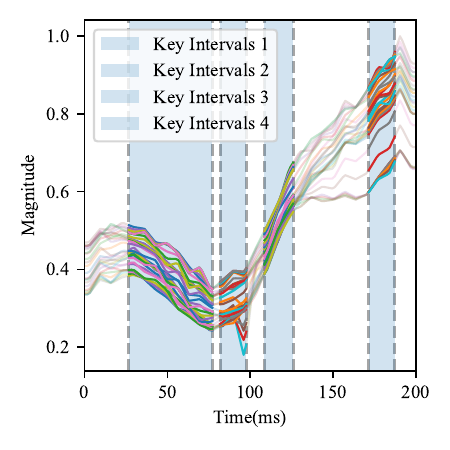}}%
  \caption{Key subcarriers and key intervals of multi-head self-attention module for different movements. (a) key subcarriers for 'Push' movement, (b) key time intervals for 'Push' movement, (c) key subcarriers for 'Bend' movement, and (d) key time intervals for 'Bend' movement.}
  \label{fig:attenScore}
\end{figure*}

Before feeding the input Frequency Token sequences into the multi-head self-attention modules, we first apply Batch Normalization for all subcarriers, where the input Frequency Token sequence \( X_f \) can be expressed as:

\begin{equation}
X_f = \text{BatchNorm}\left([F_1, F_2, \ldots, F_j, \ldots, F_{N_S}]\right)
\end{equation}

Each \(F_j\) is a vector representing the Frequency Token for the \(j\)-th subcarrier, and \(N_s\) is the total number of subcarriers. The input sequence \(X_f\) is then fed into the multi-head self-attention modules for feature extraction.

In our model, the Multi-Head Attention (MSA) is calculated by averaging the outputs of each attention head. If we denote the output of the $i$-th head as $MSA_i(y_F)$, and the result of MSA is the average of all individual head outputs:

\begin{align}
\text{MSA}(X_f) &= \frac{1}{H} \sum_{i=1}^{H} \text{MSA}_i(X_f) 
= \frac{1}{H} \sum_{i=1}^{H} \text{Softmax}\left(\frac{Q_i K_i^{\top}}{\sqrt{d_k}}\right) V_i
\end{align}
where $H$ is the number of attention heads, and each head computes query $Q_i = X_f \mathcal{W}_Q^{(i)}$, key $K_i = X_f \mathcal{W}_K^{(i)}$, and value $V_i = X_f \mathcal{W}_V^{(i)}$ through linear transformations of the input $X_f$. Here $d_k$ is the dimension of Q, K, V vectors, and $\mathcal{W}_{Q}$, $\mathcal{W}_{K}$, $\mathcal{W}_{V}$ are the weight matrices for query, key and value transformations respectively.

As shown in Fig.~\ref{fig:FeatureEX}, the attention outputs pass through a feed-forward network (FFN) that contains two linear layers with ReLU activation and dropout regularization. A residual connection sums the FFN output with its input, followed by layer normalization to produce the one-dimensional (1D) feature representation. Then, the reconstruction layer is applied to convert the 1D features into 2D feature maps for future PCM and PAF estimation.

To preserve the spatial structural relationships for human pose estimation, the reconstruction layer reshapes the 1D feature into 36×36 matrices, then uses a 3×3 convolutional layer to extract spatial patterns, followed by Batch Normalization and ReLU activation to enhance feature discriminability.

The design of time-frequency tokens enables the model to capture pose-sensitive subcarriers and key sampling moments, which makes the proposed model dynamically select features. The multi-head attention mechanism computes domain-specific attention scores to identify subcarriers temporal periods most sensitive to specific movements. This adaptive selection automatically adjusts the model's focus for different motion patterns. 

This adaptive behavior is shown in Fig.~\ref{fig:attenScore}, where we compare the model's attention patterns of "Push" and "Bend" movements. Specifically, Fig.~\ref{fig:push1} and Fig.~\ref{fig:bend1} show how pose-sensitive subcarriers dynamically change with different movements, whereas Figs.~\ref{fig:push2} and Fig.~\ref{fig:bend2} reveal the model's ability to identify and focus on key temporal intervals corresponding to distinct movements.

The attention mechanism establishes global correlations between any time-frequency points, which allows the model to efficiently perceive global information \cite{NIPS2017_3f5ee243}.

\subsection{Multi-Stage Pose Estimation}
The pose estimation consists of two parts: Pose Probability Estimator (PPE) and Pose Decoder. The PPE module processes features extracted from CSI to estimate Pose Probabilities, which consist of PCM and PAF.

The PCM is the 2D representation of the likelihood that a particular body part is located at a given pixel, whereas the PAF is the 2D representation of the likelihood that two keypoints are connected in a given direction.

The Pose Decoder determines the human pose by connecting keypoints through the interpretation of PAF obtained by PPE.

\subsubsection{Pose Probability Estimation}
We propose the Multi-Stage Feature Fusion Network (MSFN) to iteratively refine pose probability estimations.

The PCM is denoted by $\mathcal{P}_i \in \mathbb{R}^{19 \times 36 \times 36}$ for 18 anatomical keypoints (e.g., nose, shoulders, knees) and the average of 18 keypoints. The PAF is denoted by $\mathcal{A}_i \in \mathbb{R}^{38 \times 36 \times 36}$ for 19 limb connections. We denote the Pose Probability at stage \(i\) as $\text{H}_i = \{\mathcal{P}_i, \mathcal{A}_i\}$ and the encoded CSI feature at stage \(i\) as $\Phi_i \in \mathbb{R}^{256 \times 36 \times 36}$.

As shown in Fig.~\ref{fig:OVO}, the multi-stage architecture first generates an initial pose estimation through the Heatmap Decoder. The Pose-Attentive Perception Module (PAPM) then combines this initial estimation with the original CSI features to produce refined CSI features. The enhanced features are fed into subsequent Heatmap Decoder stages. This multiple stage refinement allows for the effective fusion of CSI-encoded features with pose information and enables the model to perform accurate PCM and PAF estimation in multi-person scenarios for various actions.

\begin{figure}[H]
\centering
\includegraphics[width=3.5in]{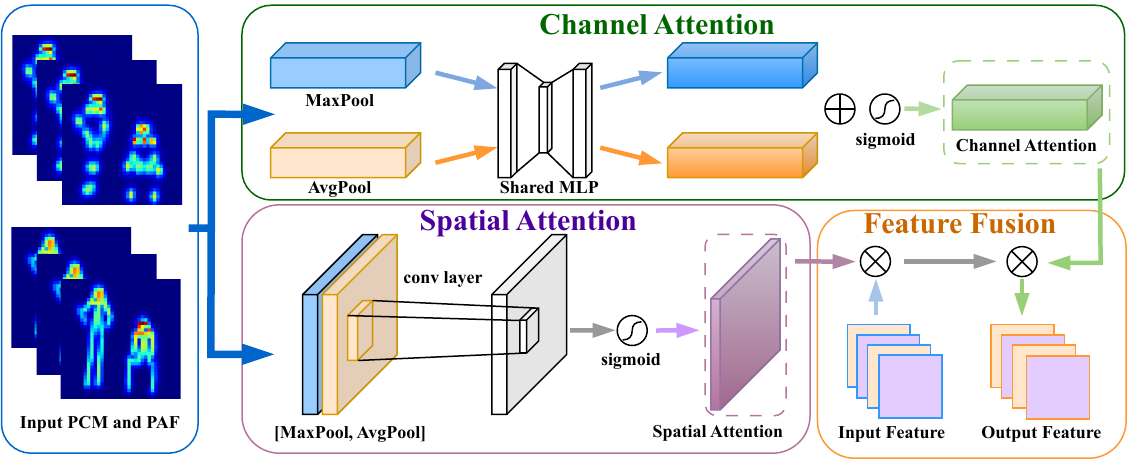}
\caption{The block diagram of Pose-Attentive Perception Module (PAPM)}
\label{fig:TwoWeights}
\end{figure}
The PAPM module is designed to enhance CSI-based pose estimation by addressing two key challenges: adaptive feature selection for multi-person scenarios and anatomical consistency in estimated poses. These challenges are addressed by the adaptive channel and spatial attention mechanisms. The channel attention mechanism adaptively identifies feature channels that are critical for pose heatmaps estimation, whereas the spatial attention mechanism emphasizes regions that contain body contours and limb connections to ensure that the estimated pose heatmaps maintain a realistic anatomical relationship. 

The feature update for stage \(i\) is formulated as:

\begin{equation}
\label{eq:feature_update}
\Phi_i = \Phi_{i-1} \odot \left( W^C_i \otimes W^S_i \right)
\end{equation}
where $W^C_i \in \mathbb{R}^{1 \times 256}$ and $W^S_i \in \mathbb{R}^{36 \times 36}$ are channel attention and spatial attention respectively, $\odot$ is the element-wise multiplication and $\otimes$ is the tensor outer product that expands both channel and spatial dimensions through broadcasting.

The channel and spatial attention weights are computed as
\begin{equation}
\label{eq:papm}
\begin{aligned}
    W^C_i &\in \mathbb{R}^{1 \times 256} = \text{PAPM}^C(PPH_{i-1}) \quad \text{(Channel Attention)} \\
    W^S_i &\in \mathbb{R}^{36 \times 36} = \text{PAPM}^S(PPH_{i-1}) \quad \text{(Spatial Attention)} \\
\end{aligned}
\end{equation}

The block diagram of channel and spatial attention mechanism in Pose-Attentive Perception Module (PAPM) is given by Fig.~\ref{fig:TwoWeights}. As shown in Fig.~\ref{fig:TwoWeights}, the spatial weights are generated from the maximum pooling and average pooling of different PAF and PCM channels. The two pooling outputs are used to generate one-dimensional channel attention weights through a shared multilayer perceptron structure. Spatial attention is obtained by convolving the pooling results of the feature maps. 

\begin{figure}[h]
\centering
\includegraphics[width=3.5in]{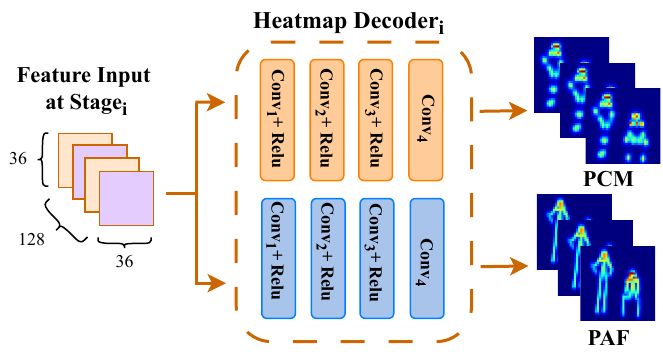}
\caption{The block diagram of Heatmap Decoder}
\label{fig:HeatmapDecoder}
\end{figure}

The Heatmap Decoder estimates PCM and PAF heatmaps from encoded CSI features by CNN to capture the spatial structure from the feature. 

\begin{equation}
\label{eq:decoder}
H_i = \{\mathcal{P}_i, \mathcal{A}_i\} =\text{Decoder}(\Phi_i; \mathbf{W}^D) 
\end{equation}

Fig.~\ref{fig:HeatmapDecoder} shows the block diagram of the Heatmap Decoder. As shown in Fig.~\ref{fig:HeatmapDecoder}, the PCM and PAF are estimated through two separate weighted CNN operations.

\begin{figure*}[!t]
  \centering
  \subfigure[\label{fig:13Pose}] 
  {\includegraphics[width=2.2in]{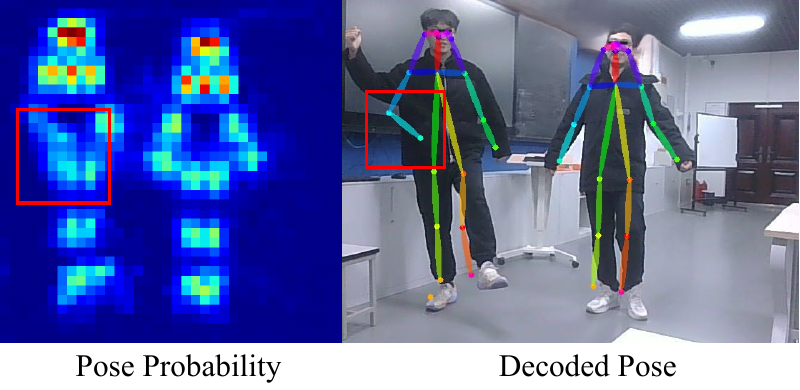}}%
  \hfill 
  \subfigure[\label{fig:23Pose}]
  {\includegraphics[width=2.2in]{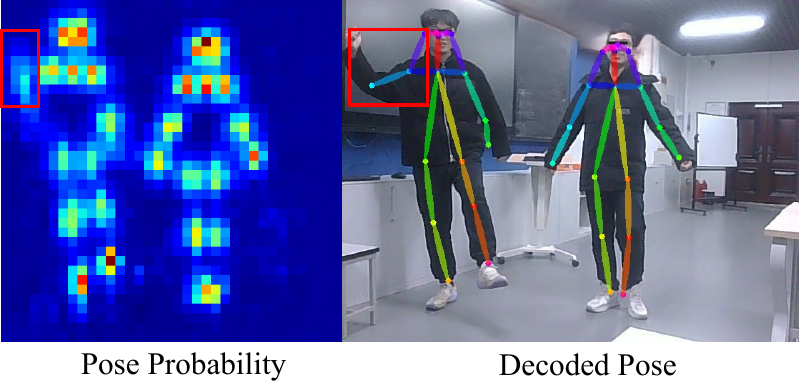}}%
  \hfill 
  \subfigure[\label{fig:33Pose}]
  {\includegraphics[width=2.2in]{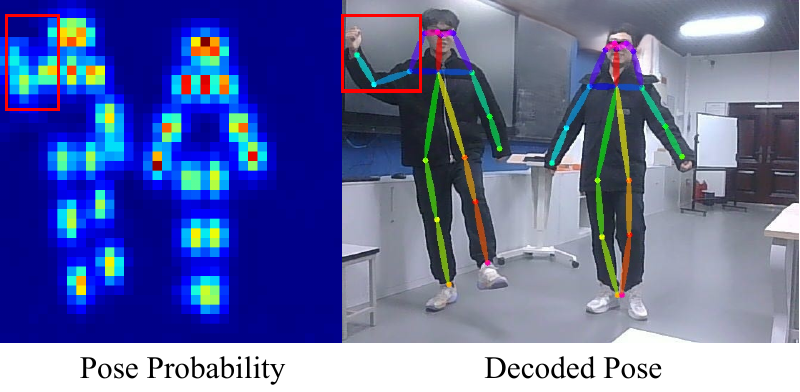}}%
  \caption{Pose probability estimation results and decoded poses of Multi-Stage Pose Feature Fusion Network (MSFN) at at different stages. (a) stage 1, (b) stage 2, (c) stage 3.} 
  \label{3stage2}
\end{figure*}

To demonstrate the effect of the iterative refinement process of MSFN, Fig.~\ref{3stage2} illustrates the estimation results of three decoding stages. As shown in this example, the MSFN learns human poses by focusing on key regions and filtering out erroneous estimations around the bodies (shown as dark shadows in the heatmaps). Specifically, Stage 1 estimates a rough skeletal outline (low-intensity parts), and in later stages, i.e., Stage 2, Stage 3, the model concentrates on the valid body regions (brightened zones). This process progressively enhances the estimation accuracy and anatomical plausibility.

\begin{figure}[H]
\centering
\includegraphics[width=3.5in]{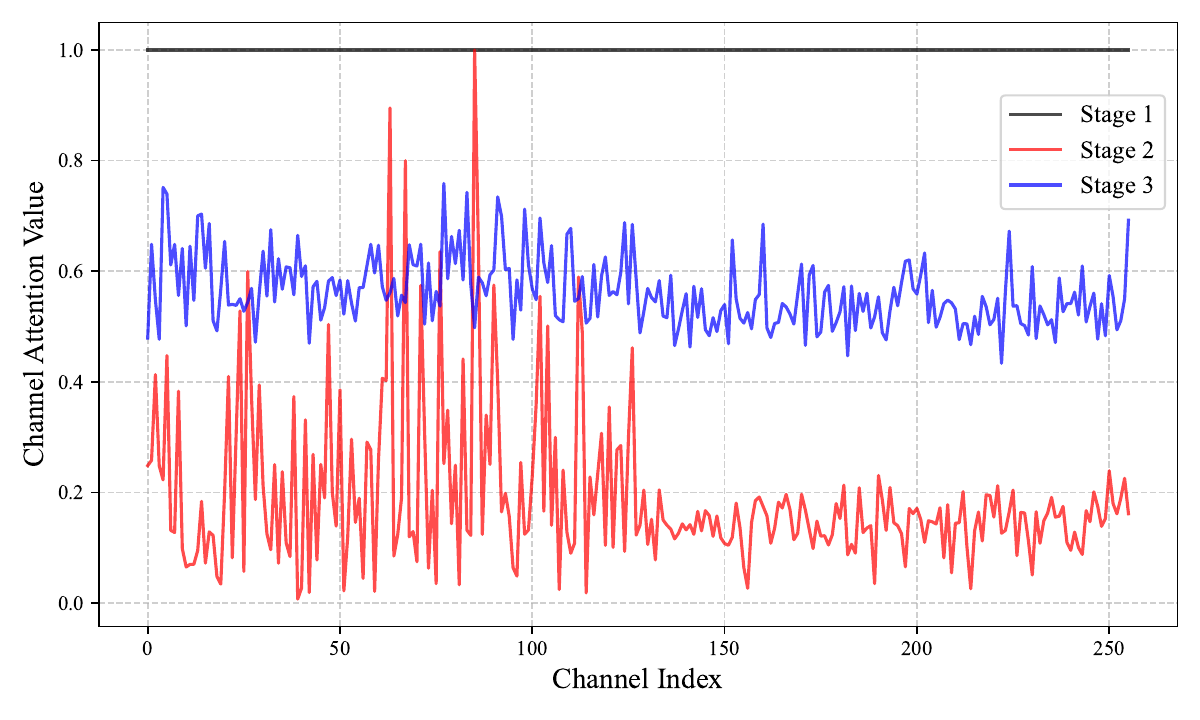}
\caption{The Channel Attention weights of Multi-Stage Pose Feature Fusion Network (MSFN) at different stages}
\label{fig:CHANw}
\end{figure}

\begin{figure}[H]
  \centering
  \subfigure[\label{fig:sa0}] 
  {\includegraphics[width=1in]{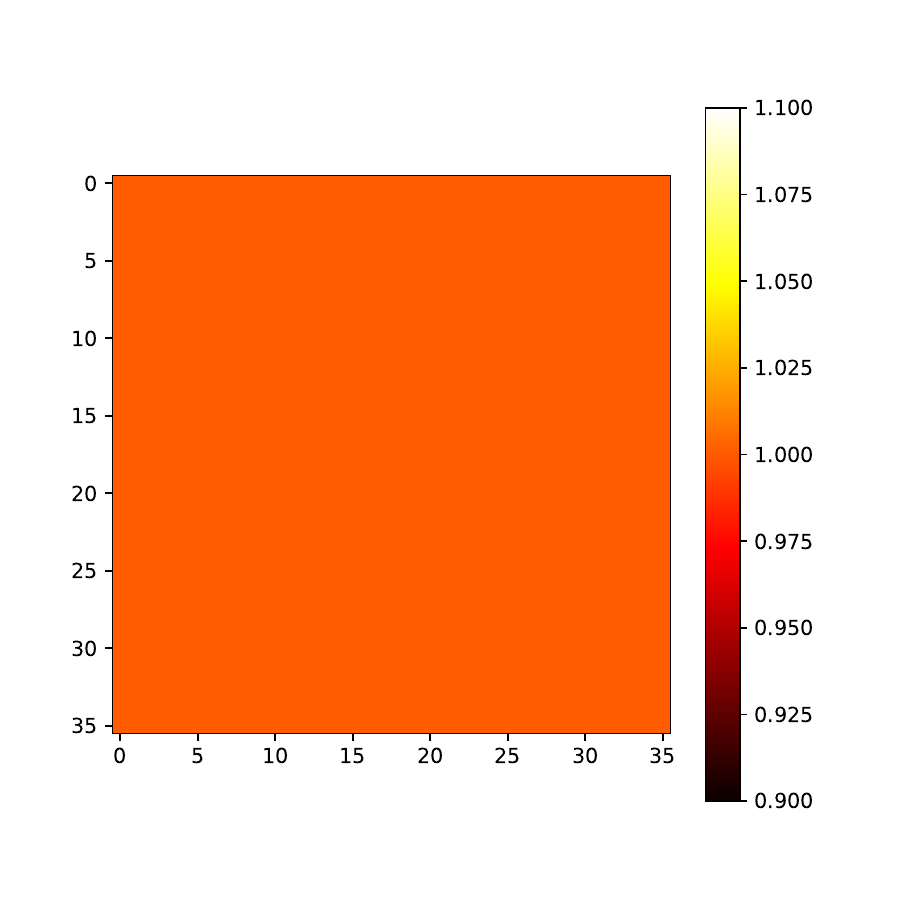}}
  \hfill 
  \subfigure[\label{fig:sa1}] 
  {\includegraphics[width=1in]{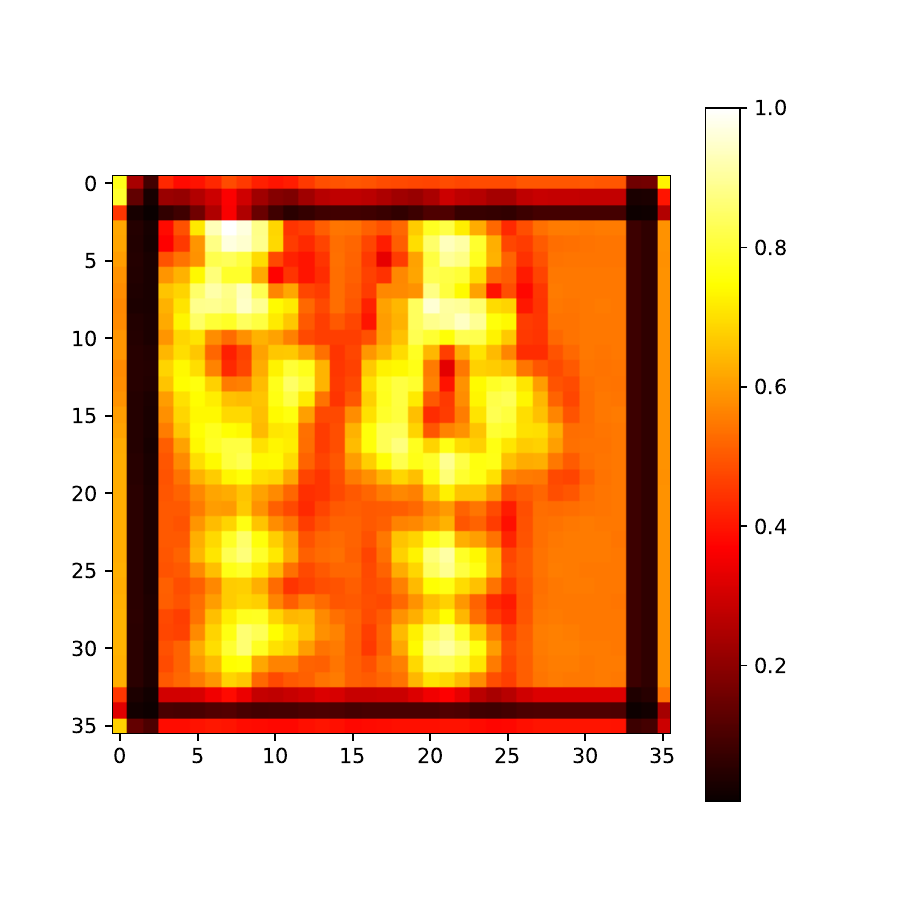}}
  \hfill 
  \subfigure[\label{fig:sa2}]
  {\includegraphics[width=1in]{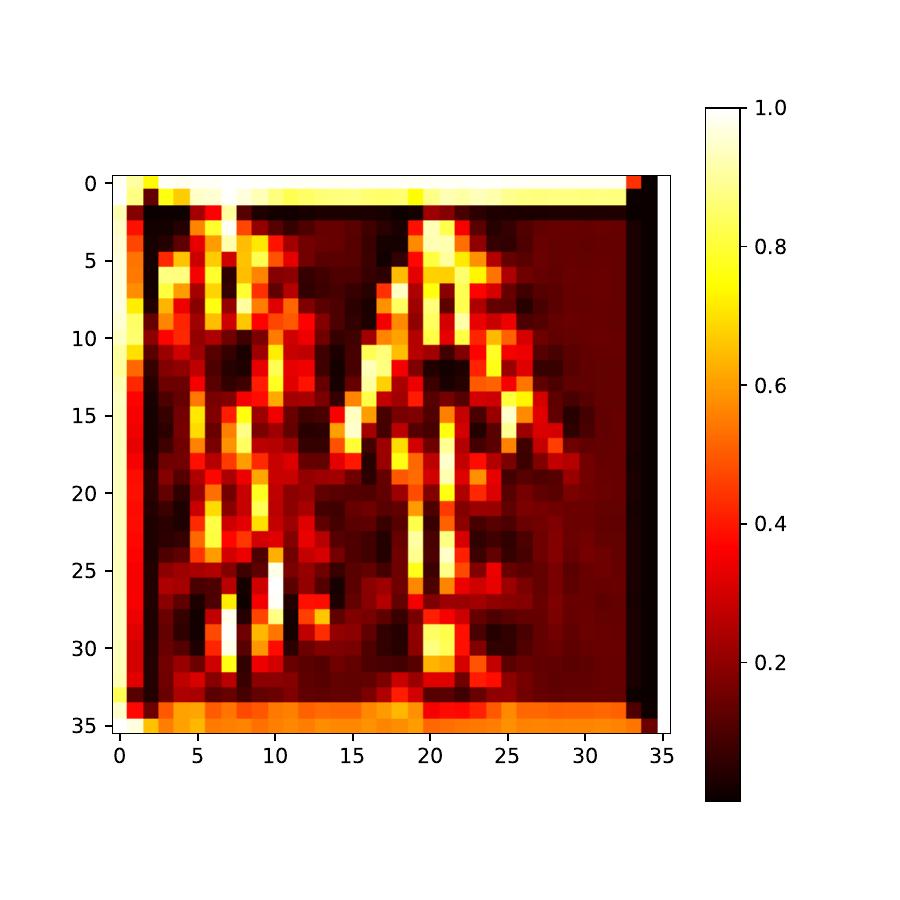}}
  \caption{Spatial Weights from Multi-Stage Pose Feature Fusion Network (MSFN) at different stages. (a) stage 1, (b) stage 2, and (c) stage 3.}
  \label{spatialweights}
\end{figure}

We show the attention weights of different decoding stages in Fig.~\ref{fig:CHANw} and Fig.~\ref{spatialweights} where all the channel and spatial weights of stage 1 are initialized to 1. From this example, we observe two phenomena: 1) Channel attention, shown by Fig.~\ref{fig:CHANw}, dynamically adjusts feature channel importance for different estimation stages to adapt to the objectives of each stage. 2) Spatial attention, shown by Fig.~\ref{spatialweights}, exhibits a progressive refinement pattern: initial stages focus on global silhouette localization, whereas the following stage shifts attention to local keypoint details.

These observations validate that the model successfully integrates CSI with pose features through its attention mechanisms, whereas adaptively optimizing its focus across different estimation stages.

\subsubsection{Pose Decoder}
After obtaining PCM and PAF from the decoder, Non-Maximum Suppression (NMS) algorithm \cite{cao_openpose_2021} is used on the PCM to obtain the candidate points for body keypoints. We denote the set of candidate keypoints by \( D_j = \{ d_j^n \} \) where \( d_j^n=\{x_j^{n_c},y_j^{n_c}\}\) is the \( n \)-th candidate of keypoint \( j \),  \( j \in \{1, \dots, 18\} \) and \( n \in \{1, \dots, N_j\} \) for maximum of \( N_j \) possible persons. We define the variable \( z_{i,j}^{m,n} \in \{0, 1\} \) to indicate whether two candidates \( d_{i}^m \) and \( d_{j}^n \) are connected. \( Z = \{z_{i,j}^{m,n}\} \) is the set of binary variables. For a pair of connectable parts, such as the left wrist and left elbow, the confidence of the connection between two keypoints can be obtained by calculating the integral of the PAF along the possible paths that connect the candidate points. The confidence of the connection between keypoints \( m \) and \( n \), denoted as \( C_{mn} \), is given by:

\begin{equation}
C_{mn} = \sum_{p \in \text{path}_{mn}} \text{PAF}(p) \cdot \frac{d_i^m - d_j^n}{\| d_i^m - d_j^n \|} 
\end{equation}
where \( p \) is a point on the path connecting \( d_m \) and \( d_n \), and \( \frac{d_n - d_m}{\| d_n - d_m \|} \) is the unit vector from keypoint \( m \) to keypoint \( n \). Our goal is to find a  set \( Z \) that ensures no two limbs of the same type share the same keypoint candidate whereas also maximizing the PAF integral \( C_{all} \). 
Formally, we define the PAF integral value \( C_k \) for the \( k \)-th type of limb connection and \(Z_k^*\) that maximizes \( C_k \) as follows:

\begin{equation}
C_k=\sum_{m \in D_i} \sum_{n \in D_j} C_{mn} \cdot z_{i,j}^{m,n}
\end{equation}
\begin{equation}
Z_k^* = \underset{Z_k}{\arg\max}~C_k
\end{equation}
subject to:
\begin{align*}
& \sum_{n \in D_j} z_{i,j}^{m,n} \leq 1, \quad \forall m \in D_i, \\
& \sum_{m \in D_i} z_{i,j}^{m,n} \leq 1, \quad \forall n \in D_j, \\
& z_{i,j}^{m,n} \in \{0, 1\}, \\
\end{align*}
where \( Z_k \) is the subset of \( Z \) corresponding to the \( k \)-th limb connection. The PAF integral \( C_{all} \) is given by \( C_{all} = \sum C_k \), and \( k \in \{1, \dots, K\} \) where \( K \) represents the number of limb types. The Hungarian algorithm \cite{cao_openpose_2021} is used to find the optimal matching \( Z_k^* \) that maximizes the PAF integral \( C_{all} \).

The Hungarian algorithm solves the keypoint association problem in multi-person scenarios by finding the optimal matching that connects candidate keypoints, while respecting anatomical constraints (e.g., left wrist to left elbow) and ensuring each keypoint is assigned to at most one person.

\section{Experiments and Result Analysis}
\subsection{Experimental Data Collection and Evaluation Metrics}

The performance of MultiFormer is evaluated on the dataset collected by us and the public dataset, i.e., MM-Fi\cite{yang2023mmfi}. The brief information of both datasets is summarized in Table \ref{tab:datasets} with detailed descriptions provided in the following subsections. 

\begin{table}[h]
  \centering
  \small 
  \caption{Datasets Comparison.}
  \label{tab:datasets}
  \setlength{\tabcolsep}{1.8pt} 
  \begin{tabular}{>{\bfseries}l c c c c c}
    \toprule
    Dataset & \multicolumn{1}{c}{Size (k)} & Volunteers & Environments & Movements & Subcarriers \\
    \midrule
    Collected & 48.7 & 2 & 2 & 8 & 30\\
    MM-Fi & 320.76 & 40 & 4 & 27 & 114\\
    \bottomrule
  \end{tabular}
\end{table}

\begin{figure}[ht]
  \centering
  \subfigure[]{\includegraphics[width=0.23\textwidth]{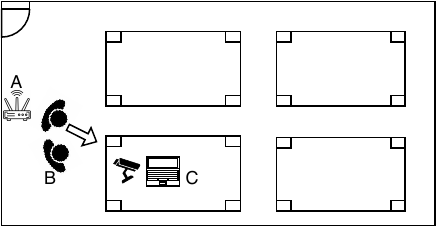}}
  \hfill 
  \subfigure[]{\includegraphics[width=0.23\textwidth]{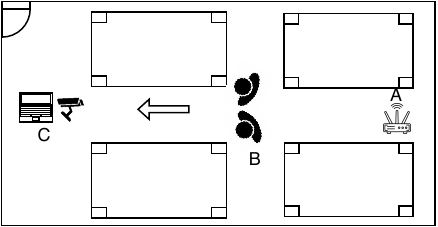}}
  \caption{Top-down view of the data collection setup. (a) Location 1, (b) Location 2.}
  \label{fig_room}
\end{figure}

We collected data in a laboratory of 9 meter $\times$ 6 meter, where we used the laptop with Intel 5300 NIC as the receiver to collect CSI. The CSI was collected at 50 Hz. We also used a webcam as the visual sensor to collect video data at the rate of approximately 8 FPS(Frames Per Second). The data collection setup is shown in Fig. \ref{fig_room}, where A is the transmitter, B are the volunteers making different movements, and C is the receiver with the webcam.

\begin{table*}[!t]
\centering
\caption{Network Architectures and Training Parameters}
\label{tab:parameters}
\renewcommand{\arraystretch}{1.5}
\begin{tabular*}{\textwidth}{@{\extracolsep{\fill}} l ccc @{}}
\toprule
\textbf{Network} & \textbf{MultiFormer} & \textbf{MultiFormer-24} & \textbf{MultiFormer-18} \\ 
\midrule

Encoder & 
8 layers, Input sequence size 64$\times$1296 & 
8 layers, Input sequence size 64$\times$576 & 
6 layers, Input sequence size 64$\times$324 \\ 

Heatmap Decoder Conv1 & 
\begin{tabular}[c]{@{}c@{}}Output Channel 128, kernel size 3, \\ Stride 1, Padding 1\end{tabular}
 & 
\begin{tabular}[c]{@{}c@{}}Output Channel 64, kernel size 3, \\ Stride 1, Padding 1\end{tabular}
 & 
\begin{tabular}[c]{@{}c@{}}Output Channel 64, kernel size 3,\\ Stride 1, Padding 1\end{tabular}
 \\ 

Heatmap Decoder Conv2 & 
\begin{tabular}[c]{@{}c@{}}Output Channel 128, kernel size 3,\\ Stride 1, Padding 1\end{tabular}
 & 
\begin{tabular}[c]{@{}c@{}}Output Channel 64, kernel size 3,\\ Stride 1, Padding 1\end{tabular}
 & 
\begin{tabular}[c]{@{}c@{}}Output Channel 32, kernel size 1,\\ Stride 1, Padding 0\end{tabular}
 \\ 

Heatmap Decoder Conv3 & 
\begin{tabular}[c]{@{}c@{}}Output Channel 512, kernel size 1,\\ Stride 1, Padding 0\end{tabular}
 & 
\begin{tabular}[c]{@{}c@{}}Output Channel 256, kernel size 1,\\ Stride 1, Padding 0\end{tabular}
 & 
\begin{tabular}[c]{@{}c@{}}Output Channel 19 (PCM) or 38 (PAF),\\ kernel size 1, Stride 1, Padding 0\end{tabular}
 \\ 

Heatmap Decoder Conv4 & 
\begin{tabular}[c]{@{}c@{}}Output Channel 19 (PCM) or 38 (PAF),\\ kernel size 1, Stride 1, Padding 0\end{tabular}
 & 
\begin{tabular}[c]{@{}c@{}}Output Channel 19 (PCM) or 38 (PAF)\\ kernel size 1, Stride 1, Padding 0\end{tabular}
 & 
--- ---\\ 
\midrule

Training Setting & \multicolumn{3}{c}{Learning rate: 1e-3, Batch size: 32, Epochs: 100, Weight decay: 0.7} \\
\bottomrule
\end{tabular*}
\end{table*}

The dataset includes the data from two volunteers performing eight different movements at two locations: bend, crouch, lean, push, sit, stand, walk, and wave. 

The MM-Fi\cite{yang2023mmfi} contains 27 human activities (14 daily actions and 13 rehabilitation exercises) performed by 40 volunteers, with synchronized CSI data and 17-keypoint pose annotations.

The network architectures and the training parameters of MultiFormer and its variants, i.e., MultiFormer-24 and MultiFormer-18 for medium and small network sizes, are given by Table \ref{tab:parameters} where the learning rate decays once every 15 epochs.

In model training, we use the Mean Squared Error (MSE) loss function to measure the difference between the model's estimated results and the ground truth. The loss function is defined as follows:

\begin{equation}
f_{pcm}^i = \sum_{j=1}^{J}\| PCM_j^i - PCM_j^* \|_2^2, 
\end{equation}

\begin{equation}
f_{paf}^i = \sum_{c=1}^{C}\| PAF_c^i - PAF_c^* \|_2^2, 
\end{equation}

\begin{equation}
f = \sum_{i=1}^{n} (f_{pcm}^i + f_{paf}^i). 
\end{equation}
where \( PCM_j^i \) is the output PCM of the \( j \)-th keypoint at the \( i \)-th stage of the MSFN, and \( PCM_j^* \) is the ground truth PCM of the \( j \)-th keypoint obtained from teacher network OpenPose\cite{cao_openpose_2021}. \( PAF_c^i \) and \( PAF_c^* \) are defined similarly for the PAF.

In the MultiFormer, we use a typical human keypoint skeleton with 18 keypoints, i.e., one nose, two eyes, two ears, one neck, two shoulders, two elbows, two wrists, two hips, two knees, and two ankles. We use the Percentage of Correct Keypoints (PCK) \cite{andriluka_2d_2014}, which is widely used in pose estimation, as our evaluation metric.

\begin{equation}
PCK_j@\alpha = \frac{1}{N} \sum_{i=1}^{N} \mathbf{I}\left( \frac{\| pd_j - gt_j \|_2^2}{\sqrt{rs^2 + lh^2}} \leq \alpha \right) 
\end{equation}where \( \mathbf{I}(\cdot) \) is the indicator function that outputs 1 when the condition is true, and 0 when it is false, \( pd_j \) is the estimated coordinate for the \( j \)-th body keypoint, and \( gt_j \) is the ground-truth value for the \( j \)-th body keypoint, \( j \in \{1, 2, \ldots, 18\} \) represents the index of the body keypoint, \( rs \) and \( lh \) are respectively the positions of the right shoulder and left hip, \( N \) is the number of tests.

We also use parameter count (model size) and FLOPs
(Floating Point Operations Per Inference) to measure the
computational efficiency of different models.

\subsection{Ablation Study}

\subsubsection{Time–Frequency Dual-Tokenization}
To evaluate the effect of the token design, we retrain the model by feeding only temporal tokens or only frequency tokens while keeping all other settings fixed.
To ensure fair comparison, the single-token models share the same structure with the dual-token models by processing identical tokens (temporal or frequency) in both branches of the feature extractor.

\begin{table}[htbp]
\centering
\setlength{\tabcolsep}{4pt}
\caption{Comparison of PCK, Parameter Count, and FLOPs for Single and Dual Token MultiFormer Models on Collected Dataset}
\label{tab:token_comparison}
\begin{tabularx}{\linewidth}{@{}lcccr@{}}
\toprule
\textbf{Feature Extractor} & \textbf{PCK@5} & \textbf{PCK@10} & \textbf{PCK@20} & \textbf{\textcolor{black}{\shortstack{Params\&FLOPs}}} \\ \midrule
Temporal Token & 0.4697 & 0.7152 & 0.8813 & \textcolor{black}{11.93M\&15.12G} \\
Frequency Token & 0.4702 & 0.7113 & 0.8502 & \textcolor{black}{11.93M\&15.12G} \\
\textbf{Dual Token} & \textbf{0.5209} & \textbf{0.7501} & \textbf{0.8885} &\textcolor{black}{ 11.93M\&15.12G} \\ 
\bottomrule
\end{tabularx}
\end{table}

\begin{table}
  \centering
  \caption{PCK@5 Comparison of 18 Body Keypoints, Parameter Count, and FLOPs from Stage 1 to Stage 3 on Collected Dataset of MultiFormer}
  \label{tab:3stage}
  \footnotesize
  \setlength{\tabcolsep}{2pt}
  \begin{tabular}{@{}lcccc@{}}
    \toprule
    \ \textbf{Body Part} & \multicolumn{4}{c}{\textbf{MultiFormer}}  \\
    \cmidrule(lr){2-5}
    & \textbf{One Stage} & \textbf{Two Stage} & \textbf{Three Stage} & \textbf{Improve (\%)}  \\
    \midrule
    Nose        & 0.3965 & 0.4268 & \textbf{0.4634} & 16.87 \\
    Neck        & 0.5376 & 0.5755 & \textbf{0.6069} & 12.89  \\
    R-Shoulder & 0.4851 & 0.5438 & \textbf{0.5845} & 20.49 \\
    R-Elbow    & 0.3626 & 0.4322 & \textbf{0.4842} & 33.54 \\
    R-Wrist    & 0.2885 & 0.3563 & \textbf{0.4268} & 47.97 \\
    L-Shoulder & 0.4757 & 0.5420 & \textbf{0.5797} & 21.86 \\
    L-Elbow    & 0.3642 & 0.4419 & \textbf{0.4967} & 36.38  \\
    L-Wrist    & 0.2343 & 0.3075 & \textbf{0.3853} & \textbf{64.41} \\
    R-Hip      & 0.4727 & 0.5576 & \textbf{0.6151} & 30.14 \\
    R-Knee     & 0.4378 & 0.5072 & \textbf{0.5533} & 26.38  \\
    R-Ankle    & 0.4035 & 0.4771 & \textbf{0.5280} & 30.85 \\
    L-Hip      & 0.4880 & 0.5578 & \textbf{0.6157} & 26.16\\
    L-Knee     & 0.4606 & 0.5150 & \textbf{0.5664} & 22.97 \\
    L-Ankle    & 0.4216 & 0.4916 & \textbf{0.5369} & 27.36  \\
    R-Eye      & 0.4046 & 0.4413 & \textbf{0.4680} & 15.68 \\
    L-Eye      & 0.4144 & 0.4403 & \textbf{0.4712} & 13.71  \\
    R-Ear      & 0.4131 & 0.4639 & \textbf{0.5001} & 21.07  \\
    L-Ear      & 0.4194 & 0.4637 & \textbf{0.4945} & 17.90  \\
   \midrule
    \textcolor{black}{\shortstack{Params\\\&FLOPs}}      & \textcolor{black}{\shortstack{1.95M\\\&2.53G}} & \textcolor{black}{\shortstack{6.85M\\\&8.82G}} & \textcolor{black}{\shortstack{11.93M\\\&15.12G}} & \textcolor{black}{\shortstack{- -\\- -}} \\
    \bottomrule
  \end{tabular}
\end{table}

The performance of \textit{MultiFormer} with different token types is compared in Table~\ref{tab:token_comparison}. \textcolor{black}{As shown, the dual-token Transformer architecture improves the PCK@5 from 0.4697 (temporal-only) and 0.4702 (frequency-only) to 0.5209, i.e.\ an average improvement of roughly 11\% , with the same parameter count of 11.93M. }\textcolor{black}{All three configurations also share the same computational cost of 15.12G FLOPs, confirming that the accuracy gain stems from dual-token modelling rather than extra computation.} This result demonstrates the dual-token model’s ability to fuse information from both temporal and frequency domains more effectively than single-token counterparts.

\begin{figure*}[htbp]
\centering
\includegraphics[width=\textwidth]{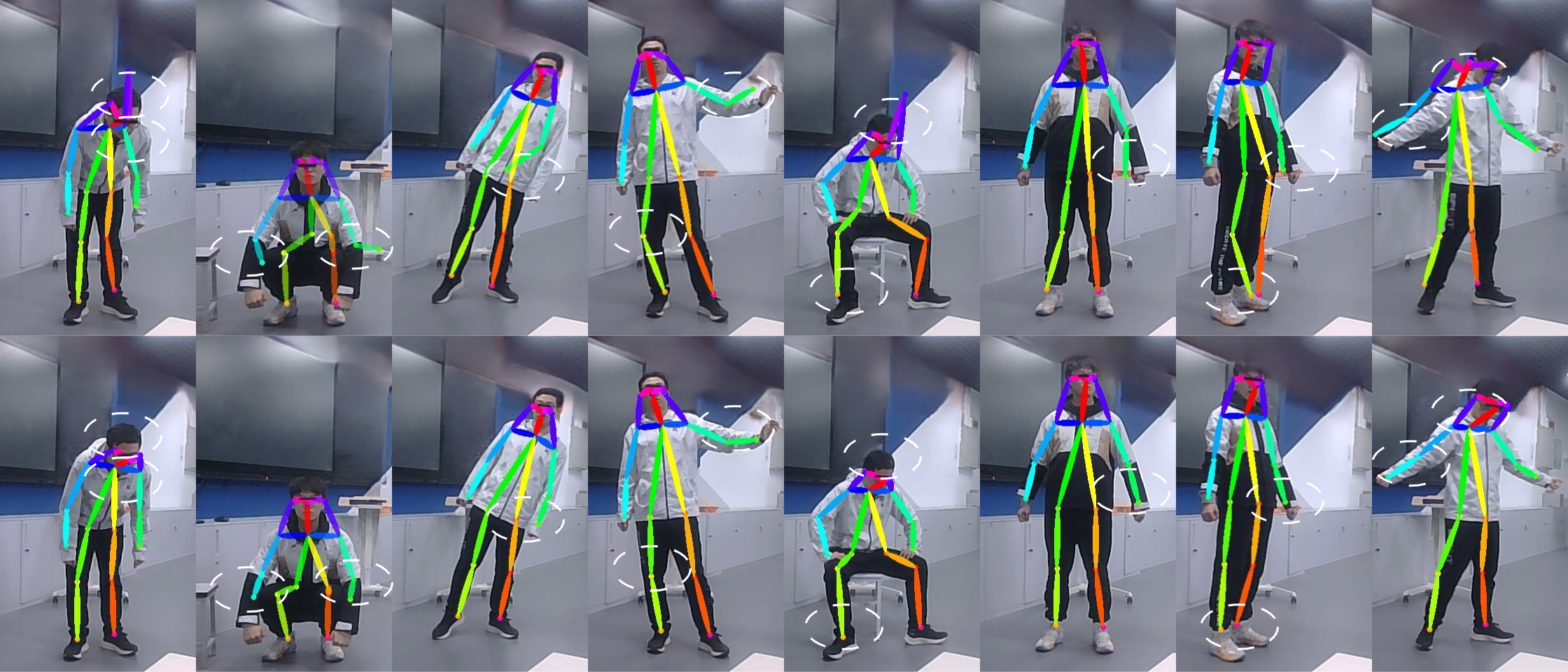}
\caption{Comparison of pose estimation results between the PAM-based method (up) and MultiFormer (down) on single-person poses for 8 movement types}
\label{fig_mat_mul}
\end{figure*}

\subsubsection{Three-Stage Feature-Fusion Decoder}
\textcolor{black}{We investigate the effectiveness of Multi-stage MSFN on the estiamtion accuracy by reducing the number of decoding stages}. In this test, Time-Frequency Dual Token-based Multi-Head Attention Network is used for feature extraction, and the estimation accuracy of different stages of MultiFormer is compared in Table \ref{tab:3stage}. It can be seen from this example, as the number of stages increases, the PCK@5 of all keypoints improves. This enhancement is particularly notable for the keypoints with larger degrees of freedom that are challenging for the one-stage estimation to estimate. Specifically, compared to the one-stage estimation, the three-stage estimation improves the PCK@5 estimation accuracy by 12\%--64\% for all keypoints, with 47\% for the right wrist and 64\% for the left wrist. This shows that MSFN has successfully learned the interrelations and constraints among human pose keypoints.

\subsubsection{Channel-Spatial Attention in PAPM}
\textcolor{black}{This experiment measures how channel-level and spatial-level attention, individually and jointly, affect the pose-estimation quality.}
\begin{table}[htbp]
\centering
\setlength{\tabcolsep}{3pt}
\caption{Comparison of PCK, Parameter Count, and FLOPs for MultiFormer Models with different PAPM on Collected Dataset}
\label{tab:papm_comparison}
\begin{tabular}{@{}lcccc@{}}
\toprule
\textbf{Feature Extractor} & \textbf{PCK@5} & \textbf{PCK@10} & \textbf{PCK@20} & \textbf{\textcolor{black}{\shortstack{Params\&FLOPs}}} \\ \midrule
Only Channel Attention & 0.3520 & 0.6269 & 0.8292 & \textcolor{black}{11.93M\&15.11G} \\
Only Spatial Attention & 0.3988 & 0.6569 & 0.8416 & \textcolor{black}{11.83M\&15.12G} \\
\textbf{Full PAPM} & \textbf{0.5209} & \textbf{0.7501} & \textbf{0.8885} &\textcolor{black}{ 11.93M\&15.12G} \\ 
\bottomrule
\end{tabular}
\end{table}

\textcolor{black}{Table~\ref{tab:papm_comparison} shows that combining both channel-level and spatial-level attention lifts PCK@5 from 0.3520 (channel-only) and 0.3988 (spatial-only) to 0.5209, corresponding to relative improvements of 48.0\% and 30.6\%, respectively.  Comparable gains appear for PCK@10 (up by 19.7\% / 14.2\%) and PCK@20 (up by 7.2\% / 5.6\%).  All three models share an almost identical computational load (11.93 M parameters, 15.1G FLOPs), confirming that the accuracy boost comes from joint channel–spatial modelling rather than extra computation.}

\subsection{Performance Comparison With Existing Methods}
 To show the effectiveness of MultiFormer, we conduct a comprehensive comparison with several well-established models, \textcolor{black}{including HPE-Li (2025)\cite{d.gian_hpeli_2025}, CKDformer (2024)\cite{YinPower2024Skel}, WPFormer (2023)\cite{zhouMetafiWiFiEnabledTransformerBased2023}, CSI-Former (2022)\cite{zhouCSIFormerPayMore2022}, CSI2Pose (2021)\cite{huang_crossmodal_2021}, and WiSPPN (2019)\cite{wangCanWiFiEstimate2019} }on our self-collected dataset.

To achieve accurate human pose estimation, it is essential to not only determine the positions of keypoints but also ensure that the estimated skeletal structure aligns with the realistic human anatomy.

 We use WISPPN, a PAM-based method, to illustrate the capability of our proposed MultiFormer to give more anatomically plausible results with the proper keypoint positioning in Fig. \ref{fig_mat_mul}. This example shows that our proposed multi-stage PCM and PAF heatmap decoder effectively models both keypoints' spatial positions and limb states, which overcomes the limitations of PAM methods that merely construct coordinate-based relational matrices to model complex human poses.

\begin{figure}[htbp]
  \centering
  \subfigure[\label{wisppn2}] 
  {\includegraphics[width=0.45\linewidth]{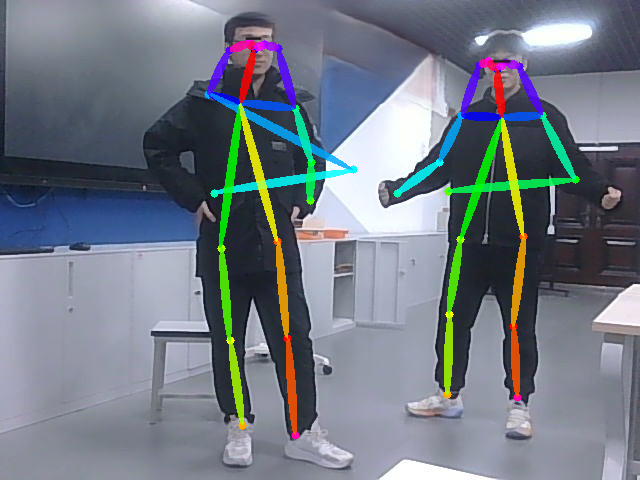}}
  \hfill 
  \subfigure[\label{MultiFormer2}]
  {\includegraphics[width=0.45\linewidth]{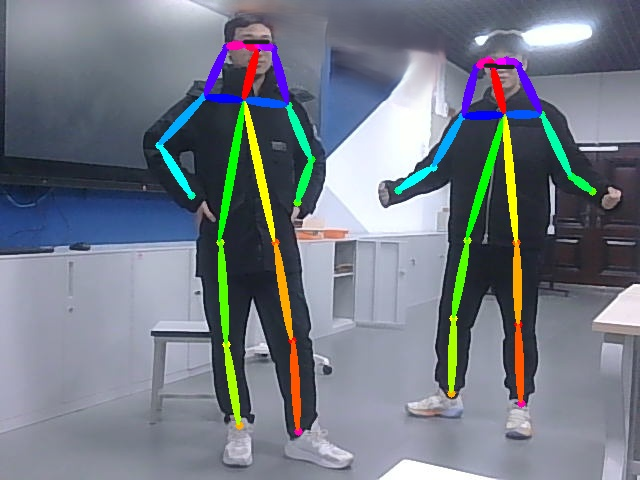}}
  \caption{Comparison of pose estimation results between the PAM-based method and MultiFormer on multi-person poses. (a) PAM-based method, (b) MultiFormer.}
  \label{fig_mulres}
\end{figure}

Another advantage of the PCM and PAF-based method is its capability for multi-person pose estimation. We show an example of multi-person pose estimation in Fig. \ref{fig_mulres}, where we compare the performance of MultiFormer with the WISPPN in the two-person setup. The PAM-based WISPPN fails to distinguish keypoints belonging to different individuals, e.g., in Fig. \ref{wisppn2}, the right elbow keypoint of the left volunteer is drawn toward the right volunteer, and the left wrist keypoint of the right volunteer is drawn toward the left volunteer, whereas the PCM and PAF-based MultiFormer avoids such misplacement in Fig. \ref{MultiFormer2}.

\begin{table}[htbp]
  \centering
  \caption{PCK@5 Comparison on Collected Dataset for 18 Body Keypoints}
  \label{tab:pck_18}
  \footnotesize
  \setlength{\tabcolsep}{0.8pt}
  \begin{tabular}{@{}lccccc@{}} 
    \toprule
    \textbf{Body Part} & \textbf{MultiFormer} & \textbf{CSI2Pose} & \textbf{\textcolor{black}{HPE-Li}} & \textbf{\textcolor{black}{CDKformer}} & \textbf{WPFormer} \\ 
    \midrule
    Nose        & \textbf{0.4634} & 0.2912 & \textcolor{black}{0.2760} & \textcolor{black}{0.2536} & 0.2534 \\ 
    Neck        & \textbf{0.6069} & 0.4256 & \textcolor{black}{0.3638} & \textcolor{black}{0.3838} & 0.3206 \\
    R-Shoulder & \textbf{0.5845} & 0.4023 & \textcolor{black}{0.3419} & \textcolor{black}{0.3491} & 0.3285 \\
    R-Elbow    & \textbf{0.4842} & 0.2910 & \textcolor{black}{0.2604} & \textcolor{black}{0.2374} & 0.2118 \\
    R-Wrist    & \textbf{0.4268} & 0.2347 & \textcolor{black}{0.2016} & \textcolor{black}{0.1622} & 0.1354 \\
    L-Shoulder & \textbf{0.5797} & 0.3788 & \textcolor{black}{0.3264} & \textcolor{black}{0.3306} & 0.2624 \\
    L-Elbow    & \textbf{0.4967} & 0.2939 & \textcolor{black}{0.2570} & \textcolor{black}{0.2385} & 0.1656 \\
    L-Wrist    & \textbf{0.3853} & 0.1984 & \textcolor{black}{0.1816} & \textcolor{black}{0.1631} & 0.1038 \\
    R-Hip      & \textbf{0.6151} & 0.4261 & \textcolor{black}{0.3772} & \textcolor{black}{0.3832} & 0.2954 \\
    R-Knee     & \textbf{0.5533} & 0.3747 & \textcolor{black}{0.3192} & \textcolor{black}{0.2960} & 0.1693 \\
    R-Ankle    & \textbf{0.5280} & 0.3324 & \textcolor{black}{0.2672} & \textcolor{black}{0.2259} & 0.0519 \\
    L-Hip      & \textbf{0.6157} & 0.4301 & \textcolor{black}{0.3766} & \textcolor{black}{0.3896} & 0.2621 \\
    L-Knee     & \textbf{0.5664} & 0.3813 & \textcolor{black}{0.3128} & \textcolor{black}{0.3095} & 0.1327 \\
    L-Ankle    & \textbf{0.5369} & 0.3357 & \textcolor{black}{0.2591} & \textcolor{black}{0.2365} & 0.1038 \\
    R-Eye      & \textbf{0.4680} & 0.2965 & \textcolor{black}{0.2801} & \textcolor{black}{0.2544} & 0.2486 \\
    L-Eye      & \textbf{0.4712} & 0.3043 & \textcolor{black}{02809} & \textcolor{black}{02603} & 0.2616 \\
    R-Ear      & \textbf{0.5001} & 0.3046 & \textcolor{black}{0.2793} & \textcolor{black}{02550} & 0.2644 \\
    L-Ear      & \textbf{0.4945} & 0.3014 & \textcolor{black}{0.2839} & \textcolor{black}{0.2805} & 0.2735 \\
    \bottomrule
  \end{tabular}
\end{table}

\begin{table}[htbp]
\centering
\caption{Comparison of the PCK@5, PCK@10, PCK@20, Parameter Count, and FLOPs for Different Models on Collected Dataset}
\label{tab:accuracy_comparison}
\begin{tabular}{@{}lcccc@{}}
\toprule
\textbf{Method} & \textbf{PCK@5} & \textbf{PCK@10} & \textbf{PCK@20} & \textbf{\textcolor{black}{\shortstack{Params\&FLOPs}}} \\
\midrule
\textbf{MultiFormer} & \textbf{0.5209} & \textbf{0.7501}   & \textbf{0.8885} & \textcolor{black}{11.93M\&15.12G} \\
MultiFormer-24    & 0.3688          & 0.6765        &0.8738  & \textcolor{black}{4.05M\&2.24G} \\
MultiFormer-18    & 0.2508          & 0.5561        &0.8176  & \textcolor{black}{\textbf{2.80M}\&\textbf{1.10G}}\\
\midrule
CSi2Pose\cite{huang_crossmodal_2021}          & 0.3335          & 0.6054        &0.8175  & \textcolor{black}{17.90M\&32.41G} \\
\textcolor{black}{HPE-Li\cite{d.gian_hpeli_2025}}          & \textcolor{black}{0.2914}          & \textcolor{black}{0.5455}        &\textcolor{black}{0.7833}   & \textcolor{black}{1.66M\&2.42G} \\
\textcolor{black}{CKDformer\cite{YinPower2024Skel}}          & \textcolor{black}{0.2783}          & \textcolor{black}{ 0.5492}        &\textcolor{black}{0.7993}   & \textcolor{black}{33.52M\&2.48G} \\
WPFormer\cite{zhouMetafiWiFiEnabledTransformerBased2023}          & 0.2136          & 0.5328        &0.8395  & \textcolor{black}{30.81M\&8.26G}  \\
CSIFormer\cite{zhouCSIFormerPayMore2022}         & 0.1868          & 0.4755        &0.7734  & \textcolor{black}{5.34M\&6.24G}  \\
WISPPN\cite{wangCanWiFiEstimate2019}           & 0.1169          & 0.3244        &0.6390  & \textcolor{black}{20.19M\&7.70G} \\
\bottomrule
\end{tabular}
\end{table}
To give a detailed comparison between MultiFormer and other models, we show the PCK@5 comparison of the 18 skeletal keypoints in Table \ref{tab:pck_18} for different pose estimation models on our self-collected dataset. In this test, MultiFormer achieves the highest estimation accuracies for all 18 keypoints, and the improvement of estimation accuracies for keypoints with high mobility is significant. For instance, MultiFormer improves the estimation accuracy for the right elbow from 0.2910 to 0.4842, which is more than 66.39\% improvement, and greater improvement can be found for the estimation accuracies of left elbows and both wrists.

Table \ref{tab:accuracy_comparison} shows the comparison of the average PCK@5, PCK@10, PCK@20 and parameter Count between MultiFormer and its lightweight variants to other models on the self-collected dataset. 

Experiment results in Table \ref{tab:accuracy_comparison} demonstrate that MultiFormer achieves the highest estimation accuracy from PCK@5 to PCK@20 \textcolor{black}{ while requiring only 15.12G FLOPs.} The estimation accuracy of the lightweight variant of MultiFormer, i.e., MultiFormer-24, achieves PCK@5 of 0.3688 \textcolor{black}{with just 2.24G FLOPs}, which outperforms other tested models of comparable parameter size. Further reducing the model to 2.8 M parameters (MultiFormer-18) \textcolor{black}{lowers the computational cost to 1.10G FLOPs }yet still delivers competitive accuracy.

\begin{table}[h]
\centering
\caption{Comparison of the PCK@20, PCK@30, PCK@40, Parameter Count, and FLOPs for Different Models on MM-Fi Dataset}
\label{tab:accuracy_comparison_mmfi}
\begin{tabular}{@{}lcccc@{}}
\toprule
\textbf{Method} & \textbf{PCK@20} & \textbf{PCK@30}& \textbf{PCK@40} & \textbf{\textcolor{black}{\shortstack{Params\&FLOPs}}} \\
\midrule
\textbf{MultiFormer} & \textbf{0.7225} & \textbf{0.8065}   & \textbf{0.8556}  & \textcolor{black}{11.93M\&15.12G} \\
MultiFormer-24      & 0.6938           & 0.7925       &0.8483    & \textcolor{black}{4.05M\&2.24G} \\
MultiFormer-18      & 0.6459           & 0.7643       &0.8304    & \textcolor{black}{2.80M\&\textbf{1.10G}} \\
\midrule
CSI2Pose\cite{huang_crossmodal_2021} & 0.6841 & 0.7304 & 0.8139 & \textcolor{black}{17.90M\&32.41G}\\
\textcolor{black}{DT-Pose\cite{chen2025towards}}     & \textcolor{black}{0.6580}        & \textcolor{black}{ 0.7790}        &\textcolor{black}{0.8510}   & \textcolor{black}{36.51M\&2.70G} \\
\textcolor{black}{CKDformer\cite{YinPower2024Skel}}     & \textcolor{black}{0.5219}       & \textcolor{black}{ 0.6843}        &\textcolor{black}{0.7813}   & \textcolor{black}{33.52M\&2.48G} \\
HPE-Li\cite{d.gian_hpeli_2025}       & 0.5207    & 0.6822  &0.7818   &  \textcolor{black}{\textbf{1.66M}\&2.42G} \\
WPFormer\cite{zhouMetafiWiFiEnabledTransformerBased2023}            & 0.4546          & 0.6444       &0.7513   & \textcolor{black}{26.42M\&507.89G} \\
WISPPN\cite{wangCanWiFiEstimate2019}              & 0.4541          & 0.6321       &0.7408   & \textcolor{black}{26.78M\&159.81G} \\

\bottomrule
\end{tabular}
\end{table}

We further evaluate MultiFormer on the MM-Fi, a public benchmark dataset\cite{yang2023mmfi}. We compare the performance of MultiFormer and its lightweight variants on the MM-Fi\cite{yang2023mmfi} with several SOTA benchmarks \textcolor{black}{including HPE-Li (2025)\cite{d.gian_hpeli_2025}, DT-Pose (2025)\cite{chen2025towards}, CKDformer (2024)\cite{YinPower2024Skel}, WPFormer (2023)\cite{zhouMetafiWiFiEnabledTransformerBased2023}, CSI2Pose (2021)\cite{huang_crossmodal_2021}, and WiSPPN (2019)\cite{wangCanWiFiEstimate2019}}
 in Table \ref{tab:accuracy_comparison_mmfi}.

As shown in Table \ref{tab:accuracy_comparison_mmfi}, MultiFormer and its lightweight variants achieve better estimation accuracy on MM-Fi from PCK@20 to PCK@40 than other tested models. The lightweight variants of MultiFormer, i.e., MultiFormer-24(4.05M) and MultiFormer-18(2.80M) achieve average 0.6938 and 0.6459 for the PCK@20, respectively, which outperform other tested lightweight model.

\begin{figure}[h]
\centering
\includegraphics[width=3.5in]{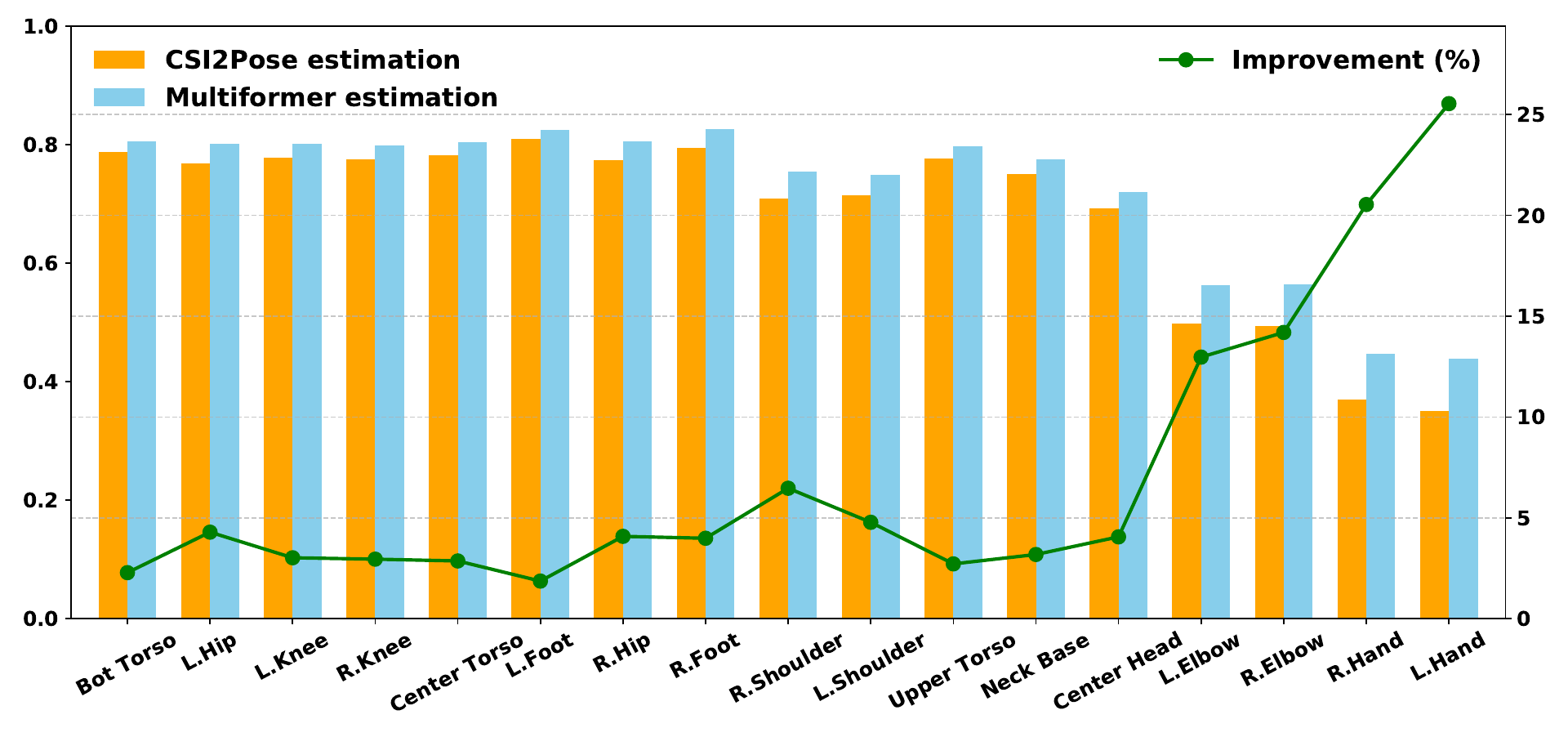}
\caption{Comparison of the PCK@20 for 17 keypoints for CSI2Pose and MultiFormer on MM-Fi datasets}
\label{fig_models2}
\end{figure}

To compare the estimation accuracy for all 17 keypoints on MM-Fi between MultiFormer and other models with similar parameter sizes, we first compare the PCK@20 of \textcolor{black}{MultiFormer (11.93 M / 15.12G FLOPs) with CSI2Pose (17.9 M / 32.41G FLOPs)}, which currently achieves the highest estimation accuracy among all tested models. The PCK@20 comparison of the 17 skeletal keypoints between MultiFormer and CSI2Pose is shown in Fig. \ref{fig_models2}. In this test, MultiFormer achieves higher estimation accuracy for all 17 keypoints \textcolor{black}{while using less than half the computation.}

\begin{figure}[h]
\centering
\includegraphics[width=3.5in]{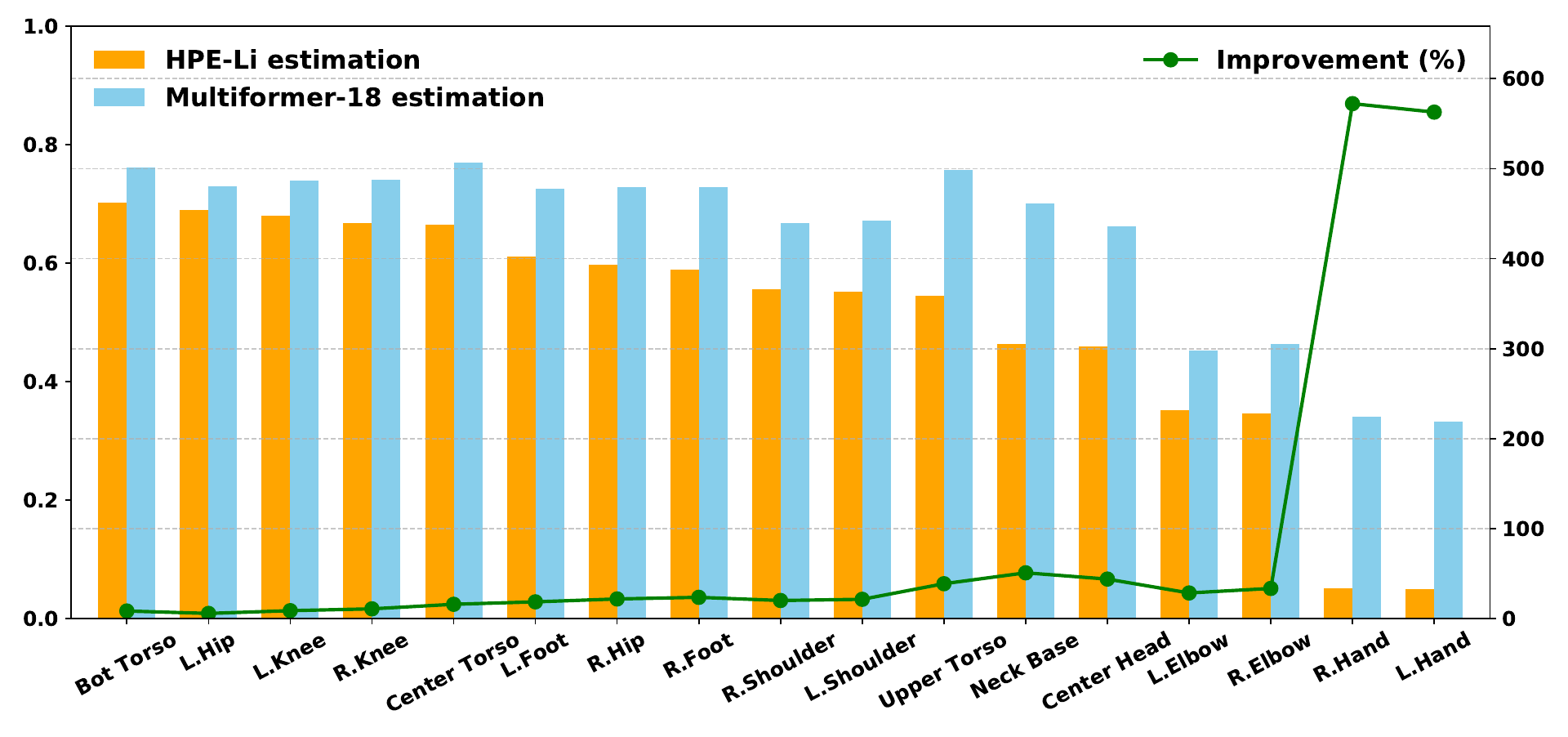}
\caption{Comparison of the PCK@20 for 17 keypoints for HPE-Li and MultiFormer-18 on MM-Fi datasets}
\label{fig_models}
\end{figure}

We also compare the PCK@20 of 17 keypoints on MM-Fi between lightweight models, i.e., MultiFormer-18(2.80M) and HPE-li(1.6M). 

The PCK@20 comparison of the 17 skeletal keypoints between \textcolor{black}{MultiFormer-18 (2.80 M / 1.10G FLOPs) and HPE-Li (1.6 M / 2.42G FLOPs)} is shown in Fig. \ref{fig_models}. MultiFormer-18 outperforms HPE-li on the MM-Fi for 17 keypoints \textcolor{black}{with roughly 55 \% less computation,} particularly for high-mobility keypoints such as wrists and elbows. Specifically, the PCK@20 for the left and right hands are 0.0503 and 0.0508 for the HPE-Li \cite{d.gian_hpeli_2025}, whereas our model achieves 0.3414 and 0.3334 with the improvement over 500\%.

\begin{figure}[htbp]
  \centering
  \subfigure[\label{black1}] 
  {\includegraphics[width=0.45\linewidth]{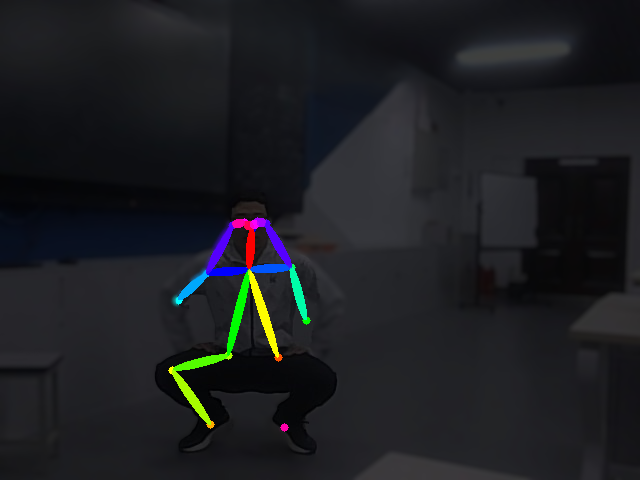}}%
  \hfill 
  \subfigure[\label{blackMultiFormer1}]
  {\includegraphics[width=0.45\linewidth]{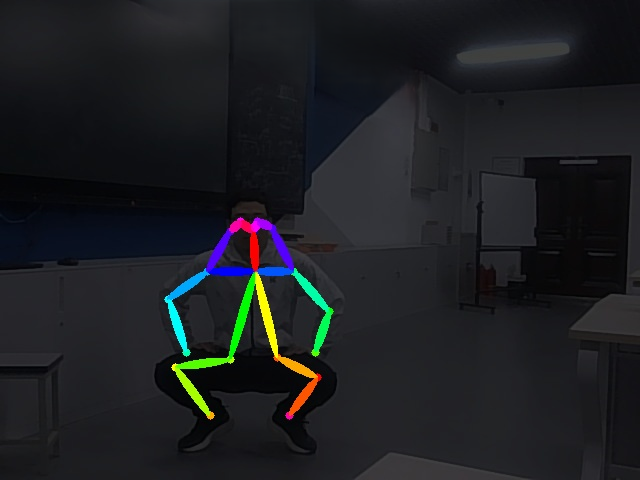}}%
  \caption{Comparison of pose estimation between OpenPose that fails to estimate PCM and MultiFormer. (a) OpenPose, (b) MultiFormer.}
  \label{fig_open_mul1}
\end{figure}
\begin{figure}[htbp]
  \centering
  \subfigure[\label{black2}] 
  {\includegraphics[width=0.45\linewidth]{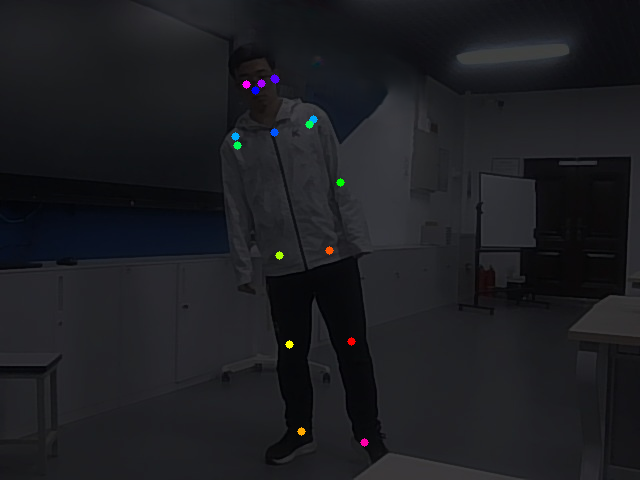}}
  \hfill 
  \subfigure[\label{blackMultiFormer2}]
  {\includegraphics[width=0.45\linewidth]{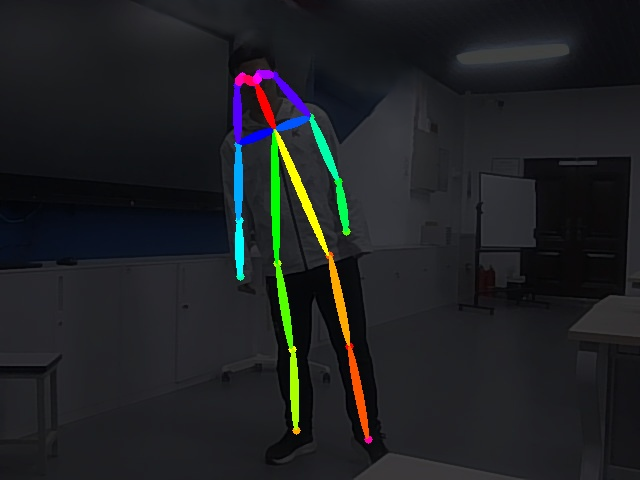}}
  \caption{Comparison of pose estimation between OpenPose that fails to estimate PAF and MultiFormer. (a) OpenPose, (b) MultiFormer.}
  \label{fig_open_mul2}
\end{figure}

We also conducted the experiment in a dark environment, and compared the pose estimation results of MultiFormer with the vision-based sensor method, i.e., OpenPose\cite{cao_openpose_2021} in Fig.~\ref{fig_open_mul1} and Fig.~\ref{fig_open_mul2}. 

In poor light conditions, vision-based sensor methods often fail to estimate the valid PCM and PAF. Fig.\ref{black1} shows a typical PCM failure case where OpenPose misses some keypoints (left/right hands and left knee), whereas Fig.~\ref{blackMultiFormer1} shows that MultiFormer is able to accurately estimate the keypoints and their connections.

Fig.\ref{black2} demonstrates the case where vision-based method is able to estimate the keypoints correctly but fail to connect these keypoints, whereas Fig.~\ref{blackMultiFormer2} show MultiFormer connects these keypoints correctly.

\section{Conclusion}

In this paper, we propose MultiFormer, a CSI-based human pose estimation system. We design the network with Time-Frequency Dual-Dimensional Tokenization (TFDDT) scheme to convert raw CSI into temporal and frequency tokens, which preserves local feature continuity in the temporal and spectral domains.The Multi-Stage Feature Fusion Network (MSFN) of MultiFormer is able to refine pose estimation through adaptive channel-spatial attention mechanisms and enforces anatomical consistency.

Experiments show that MultiFormer outperforms HPE-Li, CSI2Pose, WPFormer methods for accuracy metric, i.e., from PCK@5 to PCK@40 on the dataset we collected and the public dataset MM-Fi. The system also demonstrates robustness under poor lighting conditions and multi-person setup. 

\bibliography{jsen.bib}
\bibliographystyle{ieeetr}

\vspace{-3em} 
\begin{IEEEbiographynophoto}{Yanyi Qu}
Undergraduate student at the School of Information and Communication Engineering, University of Electronic Science and Technology of China (UESTC).
\end{IEEEbiographynophoto}
\vspace{-3em} 
\begin{IEEEbiographynophoto}{Haoyang Ma}
Undergraduate student at the School of Information and Communication Engineering, University of Electronic Science and Technology of China (UESTC).
\end{IEEEbiographynophoto}
\vspace{-3em} 
\begin{IEEEbiographynophoto}{Wenhui Xiong}
received the Ph.D. degree from Ohio University, USA in 2007. From 2007 to 2009, he was a Senior Engineer with Qualcomm Corporate R\&D Center. Since 2009, he has been an Associate Professor at UESTC's National Key Laboratory of Communications.
\end{IEEEbiographynophoto}

\end{document}